\documentclass{article}

\usepackage{subcaption}
\usepackage{todonotes}
\usepackage{booktabs}
\usepackage{changes}
\usepackage{float}
\usepackage{amsmath}
\usepackage{url}
\usepackage{authblk}
\usepackage[natbibapa]{apacite}

\begin{document}

\title{Learning inflection classes using Adaptive Resonance Theory}


\author[1]{Peter Dekker\thanks{\texttt{research@peterdekker.eu}}}
\author[1]{Heikki Rasilo\thanks{\texttt{heikki.rasilo@vub.be}}}
\author[1]{Bart de Boer\thanks{\texttt{bart@ai.vub.ac.be}}}
\affil[1]{AI Lab, Vrije Universiteit Brussel, Belgium}
\date{}


\maketitle              

\begin{abstract}
  The concept of inflection classes is an abstraction used by linguists, and provides a means to describe patterns in languages that give an analogical base for deducing previously unencountered forms. This ability is an important part of morphological acquisition and processing. We study the learnability of a system of verbal inflection classes by the individual language user by performing unsupervised clustering of lexemes into inflection classes. As a cognitively plausible and interpretable computational model, we use Adaptive Resonance Theory, a neural network with a parameter that determines the degree of generalisation (\emph{vigilance}). The model is applied to Latin, Portuguese and Estonian. The similarity of clustering to attested inflection classes varies depending on the complexity of the inflectional system. We find the best performance in a narrow region of the generalisation parameter. The learned features extracted from the model show similarity with linguistic descriptions of the inflection classes. The proposed model could be used to study change in inflection classes in the future, by including it in an agent-based model.
\end{abstract}

\section{Introduction}
In order to use language productively, speakers must be able to use words in novel contexts. In languages with inflectional morphology, this requires using the appropriate morphemes (\citealp[pp.4--5, 16--17]{blevins2017zipfian}; \citealp{ackerman2009parts}). Different words may take different morphemes in the same context, but this variability is not arbitrary. Linguists have proposed that languages can have inflection classes. An inflection class is ``a set of lexemes\footnote{A \emph{lexeme} is a lexical item, such as the verb \emph{(to) go}, which consists of multiple inflected forms – one per \emph{paradigm cell} (a person-number-tense-mood combination), such as \emph{(I) go} for first-person–singular–present (\textsc{1sg}) and \emph{(he/she/it) goes} for present \textsc{3sg}.} whose members each select the same set of inflectional realisations'' \citep[p. 64]{aronoff1994morphology}.
For example, in Latin, all lexemes in verbal inflection class I have the suffix \emph{-o} for the present \textsc{1sg} (e.g. \emph{am-o} `love-\textsc{1sg}'), while all lexemes in inflection class II have the suffix \emph{-eo} for the present \textsc{1sg}  (e.g. \emph{mon-eo} `warn-\textsc{1sg}') \citep[p. 20, Table 7]{pellegrini2019predictability}.\footnote{This example assumes an analysis where the stem is kept constant and the suffix maximised. In an alternative analysis, the stem changes across inflection classes, while the suffixes stay constant.} \footnote{According to another definition, which allows for more variation within an inflection class, inflection classes are ``classes of lexemes that share similar morphological contrasts'' \citep[p. 4]{brown2012network}.}

Inflection classes help language users to deduce unseen word forms based on the patterns characteristic to the class \citep{milin2009simultaneous,verissimo2014variables}. By analogical generalisation from other lexemes within an inflection class, the forms for unknown paradigm cells can be predicted \citep{lindsay-smith2024analogy}. Inflection classes appear to be a psychological reality for speakers \citep{enger2014reinforcement,maiden2018romance}. 
Inflection classes also play a role in language change: inflection classes can attract new words \citep{round2022cognition} and have a tendency to reinforce themselves to become more distinct from other inflection classes over time \citep{enger2014reinforcement}. Learnability of an inflectional system is seen as an important factor leading to change \citep[pp. 84--86]{elsner2019modeling}. This is especially the case for languages acquired by L2 speakers, who may have trouble learning complex morphological systems, leading to morphological simplification \citep{kusters2003linguistic,lupyan2010language}.

Earlier theoretical and computational work in morphology has focused on the task of predicting morphemes used in a paradigm cell, given knowledge of morphemes for other paradigm cells (the Paradigm Cell Filling Problem; \citealp{ackerman2009parts}) but has not modelled acquisition of inflection classes explicitly. 
Although such models may be practical for predicting morphemes for unseen contexts, they may not be good models for predicting language change. We therefore model acquisition of inflection classes explicitly. It should be noted that we focus on acquisition of regular inflection -- irregular forms that do not fall into any inflection class can by definition not be learned by our system. We assume that handling of irregular lexemes is done by the lexicon and that our model will be used to deal with lexemes whose inflection is not stored explicitly in the lexicon. In order to learn these regular inflection classes, we propose an unsupervised task of clustering lexemes into inflection classes and employ a computer model as a model of linguistic processing, to see under what circumstances a system can be learned by a language user.

Inflection class clustering is an abstraction of the processing task a language user has to perform in real communication, but the fact that prediction of linguistics features plays a central role in this task, has a basis in theories of linguistic processing.
Inflection class clustering is a specific instance of what \citet[pp. 9--10]{guzmannaranjo2019analogical} calls \emph{analogical classification}: the task for a language user to assign a certain class or feature to a lexical item, based on phonological or semantic similarity. The concept of analogy in cognition is well known (see \citealp{gentner2001analogical}), and already in 1877 Paul Hermann claims its importance in linguistics: ``Everybody who speaks continuously creates analogies” -- e.g. speech does not only reproduce patterns from experience, but new patterns are rather generated based on larger structures formed on basis of previous experiences'' (\citealp{paul1880prinzipien} as in \citealp{auer2015hermann}). Prediction of a class or feature can help a language user disambiguate which variant to use for an item (e.g. predicting the inflection class can help determine which form to use for a paradigm cell). \citet[pp. 10, 12--13, 205--206]{guzmannaranjo2019analogical} suggests that analogical classifiers, which predict classes from forms, may be able to model linguistic phenomena more naturalistically than models which predict forms from forms (such as models based on the Paradigm Cell Filling Problem discussed above), because analogical classifiers learn an intermediate linguistic abstraction, the class, which may influence linguistic behaviour. \citet{bybee2014language} corroborate with data from English that class assignment may be a process in language processing responsible for change of inflectional systems.

In the rest of this paper, we will investigate how inflection classes can be learned, using Adaptive Resonance Theory, a neural network model of category learning  \citep{carpenter1987massively}. The learning mechanism of Adaptive Resonance Theory allows us to control with a simple parameter (called \emph{vigilance}) the level of generalisation, that is: how similar do two lexemes need to behave in order to be considered members of the same inflection class. This allows us to use the appropriate level of generalisation when learning the inflectional system of a given language. Doing this is not straightforward in other neural network approaches (cf. \citealp[pp. 221--229]{goldsmith2006learning}, who need to vary the structure of their networks by hand to get appropriate generalisation). Furthermore, the model learns incrementally\footnote{The systems learns incrementally, that is by processing one item at a time. This is similar to how humans learn and in contrast with most modern neural networks that learn with batches (many items at the same time) or information-theoretic approaches that learn from the whole dataset at once.} and the features of the network are interpretable, which makes it possible to analyse which parts of words the model has used for its clustering. We evaluate the model on languages with different inflectional systems: Latin, Portuguese (both Romance, Indo-European) and Estonian (Uralic). Before we proceed to introducing the model and task setting, we will give an overview of some related work on computational methods in morphology and analogical classification.

\subsection{Computational models of morphology}
\label{sec:computationalmodels}
Morphological processing has been studied using quantitative and computational models (reviewed in \citealp{bonami2017computational}), specifically by applying sequence-to-sequence neural network models \citep{elsner2019modeling,liu2021computational}. Central in in much recent theoretical and computational work on morphology is the \emph{Paradigm Cell Filling Problem} (PCFP; \citealp{ackerman2009parts}): the task for a language user to generate an inflected form for a certain paradigm cell, given knowledge of the forms of other paradigm cells. In computational models, the PCFP has been implemented as the task of \emph{morphological reinflection}, among others in submitted systems to SIGMORPHON shared tasks \citep{cotterell2016sigmorphon,cotterell2017conllsigmorphon,cotterell2018conll,vylomova2020sigmorphon,pimentel2021sigmorphon}, which involve generalisation between typologically diverse languages \citep{kodner2022sigmorphon} and morphological reinflection in a child acquisition setting \citep{kodner2022sigmorphona}. However, this line of work does not take inflection classes into account: the inflection classes only play an implicit role in the generation of forms, instead of being explicitly modelled. For our purposes this is undesirable. If we want our models to be relevant for understanding language change, they should make the same kind of overgeneralisations and other ``errors'' in morphology as humans, and the evidence shows that inflection classes are important here \citep{round2022cognition,enger2014reinforcement, elsner2019modeling}. Moreover, PCFP models do not have to be transparent; as long as they predict the correct morphemes, their internal structures can be ignored. If one is interested in understanding language learning and how it may influence language change, it is important to be able to understand \emph{what} is learned and apply a computer model that allows for extraction of this knowledge.

\subsection{Analogical classification and inflection class clustering}
Because we are interested in inflection classes as a system, we take a different approach, which we call \emph{inflection class clustering}: to study human morphological processing, we would like to see if a computer model is able to cluster lexemes together into inflection classes, based on phonological (i.e. surface) similarity between forms, and inspect which representations it learns. The task is unsupervised: we assume the model has no access to the attested inflection classes of lexemes, the clustering is performed purely based on similarity between lexemes. 

This clustering task could be seen as an instance of \emph{analogical classification} (\citealp[pp. 9--10]{guzmannaranjo2019analogical}; \citealp[p. 225]{guzmannaranjo2020analogy}): class assignment to a lexical item based on similarity between the items. Analogical classifiers have been proposed for the prediction of several class systems or features, such as lexical gender \citep{eddington2002spanish,matthews2005french,matthews2010nature}, competition between suffixes \citep{arndt-lappe2014analogy}, stress assignment in compounds \citep{arndt-lappe2011exemplarbased}, prediction of etymological origin of nouns in Maltese \citep{court2023analogy} and predicting whether verbs belong to the regular or irregular class \citep{bybee1982rules,matthews2013analogical}.

For the prediction of inflection classes, a number of supervised approaches have been proposed, where, during training, the model is provided with information about the inflection classes derived by linguists. Information-theoretic \citep{beniamine2018inferring,lefevre2021formalizing}, decision tree-based \citep{bonami2022derivation} and neural network approaches \citep{guzmannaranjo2019analogical,guzmannaranjo2020analogy,williams2020predicting} have been applied to supervised inflection class prediction for noun and verb systems in different languages, such as French, Spanish, German and Czech.

As we assume real language users do not get told explicitly which inflection class a word belongs to, but rather rely on more implicit information, unsupervised clustering of lexemes, where no inflection class labels are provided, seems a more plausible task.
Unsupervised inflection class clustering approaches on text corpora include \citet{monson2004unsupervised}, who infer inflection classes from a Spanish news corpus, by searching for patterns of stems and suffixes in the data, and \citet{eskander2013automatic}, inferring inflection classes from morphologically annotated corpora of Egyptian Arabic and German.
\citet{brown2012morphological} use a compression-based unsupervised machine learning algorithm to cluster Russian nouns. \citet{lee2014automatic} apply an algorithm using Minimum Description Length to group verbs into inflection classes in an unsupervised fashion for English and Spanish verbs. The models of
\citet{beniamine2018inferring} and \citet{lefevre2021formalizing} could be seen as partially unsupervised. They use information on finer-grained microclasses to infer coarser-grained macroclasses using Minimum Description Length. Even though microclasses can in theory be inferred from the data (see Section \ref{sec:datasets}), this requires additional steps of computation.

What we propose in this paper is the clustering of inflection classes, based on phonological similarity between forms, in an unsupervised way. To perform this task, we need a model with cognitive plausibility, which provides interpretability of the learned features of the model.

\subsection{Adaptive Resonance Theory}
\label{sec:art-background}
\citet[pp. 24, 29--30, 33]{guzmannaranjo2019analogical} suggests, after \citet[p. 81]{matthews2005french}, that when predicting inflection classes, there is no fundamental difference between the use of different model architectures, such as neural networks or Analogical Modelling, as long as they are applied in the same setting of a classification task. Even though from the perspective of the performed task (the \emph{computational level} in \citealp{marr1982vision}'s tri-level hypothesis), any sufficiently powerful model would work, if the goal is to better understand the cognitive processes involved in the task, it is crucial to use a model architecture where the processes and representations (Marr's \emph{algorithmic} level) are interpretable in cognitive terms. 
\citet[pp. 221--229]{goldsmith2006learning} suggest generalisation, from known forms to unseen forms as a key mechanism behind the formation of inflection classes and propose that neural network models may be a good way to model generalisation. Furthermore, as real language users encounter linguistic input partially and incrementally \citep{ackerman2009parts,blevins2017zipfian}, it would be good to use a model that learns incrementally from data. We therefore adopt a neural network model that includes generalisation as an explicit parameter and learns in an incremental fashion.

In our approach, we use Adaptive Resonance Theory 1 (ART1; \citealp{carpenter1987massively}), which the authors present as a cognitively inspired neural network of category learning. It clusters binary vectors in an unsupervised fashion. ART1 is an instance of \emph{Adaptive Resonance Theory} (ART, \citealp{grossberg1976adaptive,grossberg1976adaptivea,grossberg1980how,grossberg2013adaptive}): a theory of attention, learning and recognition in the brain and at the same time a family of neural network models (see \citealp{britodasilva2019survey} for an overview of model architectures). This family includes unsupervised (such as ART1 for binary and FuzzyART for real-valued inputs, \citealp{carpenter1991fuzzy}), supervised (such as ARTMAP, \citealp{carpenter1991artmap}) and reinforcement learning models (e.g. TD-FALCON, \citealp{tan2008integrating}). ART models have been practically applied in engineering tasks \citep{caudell1994nirs,jain2000innovations,lerner2008advanced}, but have also been used as cognitive models to study aspects of language, such as the cognitive lexicon \citep{dunbar2012adaptive} and implicit learning of orthographic word forms \citep{glotin2010adaptive}.

\begin{figure}
    \includegraphics[width=\linewidth]{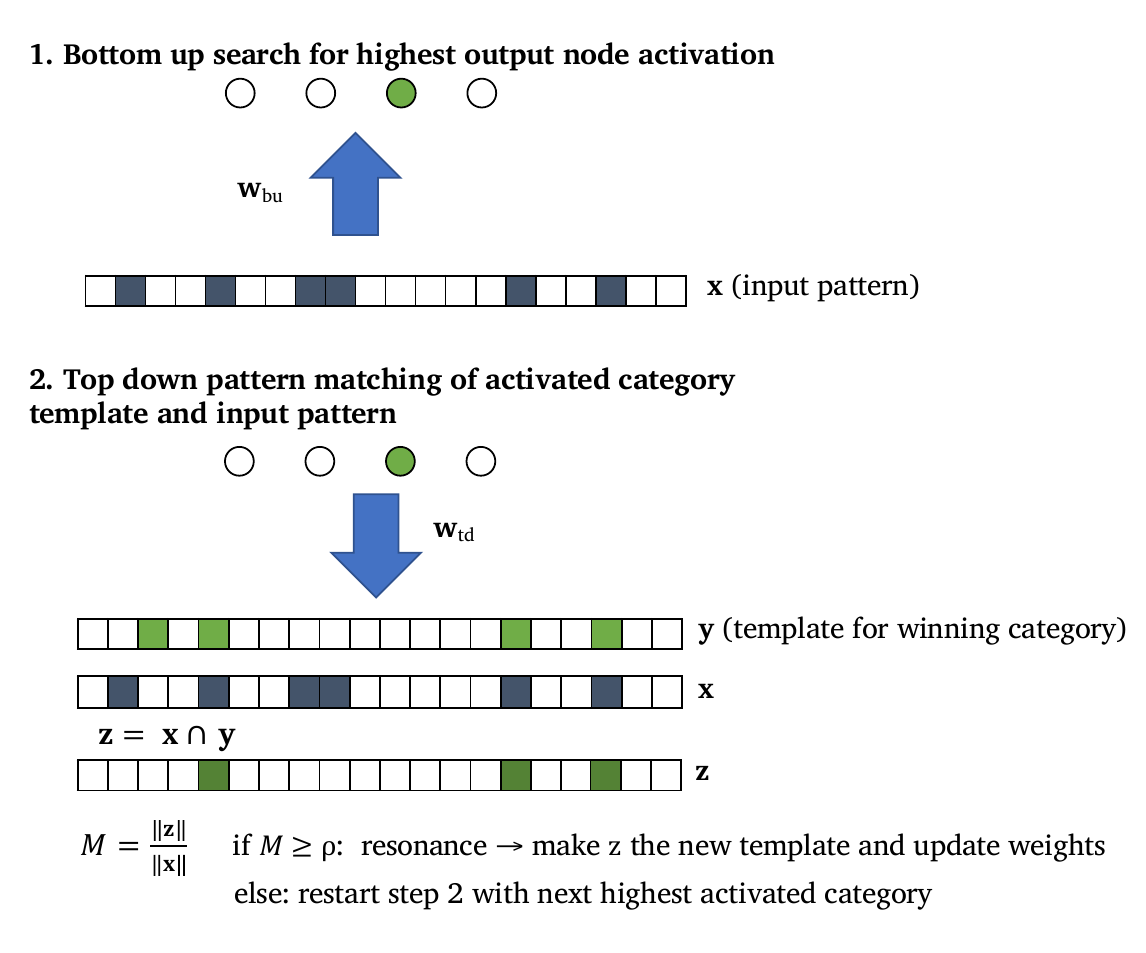} 
    \caption{Illustration of the phases of training of the ART1 network. Step 1) Input sample $\boldsymbol{x}$ is propagated through the bottom-up weights, and the output node with the highest activation is selected as the hypothesised category for this sample. Step 2) Top-down weights for this category are used to access the category template $\boldsymbol{y}$. Logical \textsc{and} operation leads to the shared feature vector $\boldsymbol{z}$ between the template and the input. If the match $M$ between $\boldsymbol{z}$ and $\boldsymbol{x}$ is higher or equal than the vigilance value $\rho$, resonance occurs and the category template is updated to match $\boldsymbol{z}$, and the bottom-up weights are updated accordingly. If vigilance value is not reached, the search restarts from the category with the next highest activation.}
    \label{fig:ART1-illustration}
\end{figure}

Although ART has not been more widely used in linguistics tasks, its characteristics make it particularly interesting for the task in hand. ART explicitly models the interaction between new observations (bottom-up) and existing knowledge (top-down). ART models are based on \emph{unsupervised, competitive learning} which Grossberg has proposed as more cognitively plausible than error-based learning mechanisms such as backpropagation, used in many neural network models \citep{grossberg2020path}. ART is more cognitively plausible because it does not depend on a global error function, but bases its updates on local interactions between nodes representing categories. ART allows for incremental learning, processing datapoints one by one, and immediately incorporates the full information of a new datapoint into the model\footnote{\citet{grossberg2020path} calls this \emph{fast learning}, as opposed to \emph{slow learning} where gradual weight updates are made based on the error signal.}, while at the same time avoiding \emph{catastrophic forgetting} of existing knowledge through the use of top-down information \citep{grossberg2013adaptive}. Central to ART models in general, and ART1 in particular, is the \emph{vigilance} parameter, controlling the degree of generalisation. If the vigilance parameter is low, a new input sample is more likely to be added to an existing category, while if it is high, it is more likely that a new category will be created.

ART1 consists of two layers: an input layer $F_1$ and an output layer $F_2$. These are connected by two sets of weights: bottom-up and top-down weights. In ART1, training and inference happen at the same time. Figure \ref{fig:ART1-illustration} illustrates the training and inference procedure of the ART1 network.\footnote{The formulae in Figure \ref{fig:ART1-illustration} are based on the description in \citet[pp. 4--5]{britodasilva2019survey}.}

Datapoints, which are binary feature vectors, are processed one by one and inputted in input layer $F_1$. The initial bottom-up activation $T_j$ of a category $j$ is calculated by taking the inner product between the datapoint $\boldsymbol{x}$ and the bottom-up weights $\boldsymbol{w}_j^{bu}$.
\begin{equation}
T_j = \langle \boldsymbol{w}_j^{bu}, \boldsymbol{x} \rangle
\end{equation}

Then, the category $J$ in the output layer $F_2$ is selected that has the highest activation $T_j$:
\begin{equation}
J = \mathrm{arg\,max}(T_j)
\end{equation}

This category, that is the winner based on the bottom-up weights, is now treated as a hypothesis. Now, the input will be compared to the top-down weights that are specific to this category. The top-down weights for this category, $\boldsymbol{w}_J^{td}$, form the \emph{template}, representing features that this category attends to. An intersection (logical \textsc{and}) between the input and template is calculated:

\begin{equation}
    \boldsymbol{z_{x,J}} = \boldsymbol{x} \cap \boldsymbol{w}_J^{td}
\end{equation}

A match $M_j$ between the $L_1$ norms of the intersection $\boldsymbol{z}_{x,J}$ and the original datapoint is calculated. This represents how many features in the input match the category template:
\begin{equation}
    M_J = \frac{\lVert \boldsymbol{z}_{x,J} \rVert_1}{\lVert \boldsymbol{x} \rVert_1}
\end{equation}

If the input pattern and the category template are similar enough according to the vigilance $\rho$ ($M_J \ge \rho$), the datapoint is assigned to this category. The bottom-up and top-down weights for this category are then updated. The top-down weights for this category are set to the intersection $\boldsymbol{z}_{x,J}$, this is thus again a binary vector:
\begin{equation}
    \boldsymbol{w}_J^{td} = \boldsymbol{z_{x,J}}
\end{equation}
The bottom-up weights are based on the updated top-down weights, but normalised by the $L_1$ norm of the top-down weights and mediated by the parameter $L > 1$. The bottom-up weights are real-valued:
\begin{equation}
    \boldsymbol{w}_J^{bu} = \frac{L}{L-1 + \lVert \boldsymbol{w}_J^{td} \rVert_1} \boldsymbol{w}_J^{td}
\end{equation}

If the vigilance value is not overcome, the search procedure restarts, excluding the categories that were tested as hypotheses earlier. This is done by finding the next highest activated category using the bottom-up weights, and repeating the top-down matching procedure. This search proceeds until the comparison of a template and the input pattern overcomes the vigilance value or all categories have been tested. In that last case, a new category is created. 

Note that the bottom-up weights are based on the top-down weights, but normalised by the norm of the top-down weights, which represents how many features are active. This makes it possible that one category has the highest bottom-up activation for a datapoint, but the datapoint eventually gets assigned to another category that has a higher match based on the top-down weights.\footnote{A datapoint is not necessarily assigned to the category with the highest top-down weights either: the datapoint may be assigned to a category processed earlier, for which the match overcomes the vigilance value.} The categories can be differently ordered in their magnitudes in the top-down and bottom-up weights, based on the number of features that matched (captured by the norm) for previous assigned datapoints. 
Intuitively: ART1 results in a clustering that is a compromise between how precisely the clusters match with the input and the number of clusters; this is modulated by the vigilance parameter.

ART1 is a competitive learning model, involving a winner-takes-all strategy: one output node is chosen as the winning category for any input sample and only the weights connected to this node will be updated (which contrasts with backpropagation approaches, where all weights are updated).

It is also inherently interpretable: the top-down weights directly represent category \emph{templates}: the features that a certain category attends to \citep{grossberg2020path}. When a datapoint is assigned to a category, the new top-down weights are updated by performing a logical \textsc{and} between the datapoint and the current top-down weights. The model thus learns the full information from the datapoint (relative to the information contained in the category) upon the first time it is presented. The template for a category, one column in the top-down weights, represents an intersection between all assigned datapoints: only those features are activated that occur in all datapoints assigned to that category. For our task, the template provides immediate access to the trigrams (see section \ref{sub:encoding}) that characterise an inflection class. 

\section{Method}
\subsection{Datasets}
\label{sec:datasets}
We use datasets of phonetic verb forms, attested with inflection classes for three languages, with some genealogical and typological variation: Latin, Portuguese and Estonian. Portuguese and Latin belong to the Italic branch of the Indo-European language family. Estonian is a Uralic language from the Finnic branch. In addition to the diversity of the morphological systems, an additional consideration to choose these three languages was the availability of datasets of phonetic forms attested with inflection classes.

As we study how a computer model can infer the inflection class, we need a dataset that contains the inflected forms for different person-number-tense-mood combinations (paradigm cells), rather than just the lemma form for that verb. In addition, we need a dataset where verbs are annotated with inflection classes for evaluation. \citet[pp. 469, 490, 512]{beniamine2018inferring}, following \citet{dressler1987leitmotifs} and \citet{dressler1996italian}, propose a distinction between \emph{microclasses}, groups of lexemes that are inflected in exactly the same way, and \emph{macroclasses}, a higher-level grouping of lexemes with similar inflections. There is not one agreed-upon way to arrive at macroclasses, they are sometimes based on post-hoc criteria or defined for pedagogical purposes \citep[pp. 476]{beniamine2018inferring}. Both microclasses and macroclasses have cognitive validity. Microclasses are directly observable, while the role that inflection classes play in language change \citep{round2022cognition,enger2014reinforcement} implies that lexemes are not always assigned to a class that has exact morphological identity, there must also be macroclass-like representations. For Latin, we use macroclasses as a first test case, while for Portuguese and Estonian we use microclasses.

For Latin, we used version 2.0.4 of the \emph{Romance Verbal Inflection Dataset} 2.0 \citep{beniamine2020opening}, a dataset of phonetic word forms for 74 Romance languages. This dataset is a republished version of the Oxford Online Database of Romance Verb Morphology, cleaned up and published in the CLDF data format \citep{forkel2018crosslinguistic}. We use the Latin portion of this dataset -- as all languages are annotated with the inflection classes of the Latin proto-forms, the inflection classes most reliably describe the verbal system for Latin. The dataset is annotated with macro inflection classes. In Latin, the inflection classes only describe the variation of affixes in the so-called \emph{infectum}, the non-perfective tenses, including the present and the imperfect, but excluding perfective tenses like the perfect and pluperfect. All forms in the infectum are based on the so-called \emph{first stem}, and membership of an inflection class determines how the subject marking suffixes differ. For other cells, which are based on the second or third stem, subject markers are always the same. Latin has 5 macroclasses: 4 classes characterised by the \emph{theme vowel} in the stem of the verb and one class \emph{special}, which contains irregular verbs which are qualitatively different from each other. Four lexemes from the dataset were omitted, because not all paradigm cells (see Section \ref{sec:selection}) were available for these lexemes. We preprocessed the data by removing defective forms (no form exists for this paradigm cell in this language), which are represented by a placeholder \texttt{Ø} in the dataset. 
Subsequently, the data is split into tokens (merging phonemes with their diacritics) using the \texttt{ipa2tokens} method from \texttt{lingpy} \citep{list2019lingpy}.

We use \emph{European Portuguese Verbal Paradigms in Phonemic Notation} for Portuguese \citep{beniamine2021fine,perdigao2021european} and \emph{Eesthetic} for Estonian \citep{beniamine2024eesthetic,beniamine2024eesthetica}, which are both published in the \emph{Paralex} standard for morphological lexicons \citep{beniamine2023paralex}\footnote{\url{https://paralex-standard.org/}}. The data is already tokenised into segments. These datasets are annotated with finer-grained microclasses, which describe the real variation in verb paradigms better.

In Portuguese, inflection classes describe affix variation and the three major inflection classes are characterised by their theme vowels like in Latin. However, these macroclasses are further split into smaller microclasses, which differ among others in their stress and vowel alternations \citep[p. 5]{beniamine2021fine}. In this Portuguese dataset, every class has the name of a verb representing that class, such as \emph{beber} or \emph{amar}.

In Estonian, inflection classes do not determine the variation of the subject marking suffixes, which are the same in all inflection classes. Instead, inflection classes determine variation in the stem: every inflection class has a characteristic pattern of cells that are in weak or strong grade \citep{blevins2007conjugation}. This gradation is a morphophonological phenomenon and affects the consonants of the stem, where compared to the strong variant, the weak variant is shorter or has undergone a process of deletion, assimilation or devoicing \citep{trosterud2005consonant,bakro-nagy2022consonant}. The cells in a paradigm can be structured into \emph{series}: cells that share the same grade within an inflection class. Table \ref{tbl:estonianclasses-blevins2007} shows how macroclass I has strong grade for all cells in the infinitive series and weak grade for the present series, while class II has the inverse pattern: weak infinitive series and strong present series. The supine series is always strong and the impersonal series always weak. For class III, all verb forms are invariant: they are all in weak grade or in strong grade, there is no alternation within the paradigm. The three macroclasses are thus defined by their strong-weak alternation. These classes are further subdivided in microclasses based on morphologically and phonologically conditioned variation, with the strong-weak alternation staying intact \citep[p. 255]{blevins2007conjugation}.

\begin{table}
\begin{tabular}{llllll}
\toprule
Series&Form&I&II&III&\\
\midrule
\textbf{Infinitive}&\textsc{inf}&\textbf{õppida}&\textit{hüpata}&elada&tarbida\\
&\textsc{imp.prs.2pl}&\textbf{õppige}&\textit{hüpake}&elage&tarbige\\
&\textsc{imp.prs.pers}&\textbf{õppigu}&\textit{hüpaku}&elagu&tarbigu\\
&\textsc{ger}&\textbf{õppides}&\textit{hüpates}&elades&tarbides\\
&\textsc{ptcp.pst.pers}&\textbf{õppinud}&\textit{hüpanud}&elanud&tarbinud\\
\midrule
\textbf{Present}&\textsc{ind.prs.1sg}&\textit{õpin}&\textbf{hüppan}&elan&tarbin\\
&\textsc{cond.prs.pers}&\textit{õpiks}&\textbf{hüppaks}&elaks&tarbiks\\
&\textsc{imp.prs.2sg}&\textit{õpi}&\textbf{hüppa}&ela&tarbi\\
\bottomrule
\textbf{Supine}&\textsc{sup}&\textbf{õppima}&\textbf{hüppama}&elama&tarbima\\
&\textsc{ptcp.prs.pers}&\textbf{õppiv}&\textbf{hüppav}&elav&tarbiv\\
&\textsc{quot.prs.pers}&\textbf{õppivat}&\textbf{hüppavat}&elavat&tarbivat\\
&\textsc{ind.pst.ipfv.1sg}&\textbf{õppisin}&\textbf{hüppasin}&elasin&tarbisin\\
\midrule
\textbf{Impersonal}&\textsc{ind.prs.impers}&\textit{õpitakse}&\textit{hüpatakse}&elatakse&tarbitakse\\
&\textsc{ind.pst.ipfv.impers}&\textit{õpiti}&\textit{hüpati}&elati&tarbiti\\
&&\ldots&\ldots&\ldots&\ldots\\
\bottomrule
\end{tabular}
\caption{Verbal inflection classes in Estonian are characterised by the distribution of strong and weak grade forms across the paradigm. Bold is strong grade, italic is weak grade, normal typeface is invariant. Table based on Table 4 from \citet{blevins2007conjugation}, after \citet[p. 52]{viks1992vaike}.}
\label{tbl:estonianclasses-blevins2007}
\end{table}

Table \ref{tbl:datasets} shows for every dataset the number of lexemes (each consisting of a number of paradigm cells) and inflection classes of the portion of the dataset we used. This table shows that the dataset for Latin is much smaller than the other two datasets and is annotated with coarser-grained inflection classes.

\begin{table}
\begin{tabular}{lll}
\toprule
Language&Lexemes&Inflection classes\\
\midrule
Latin&227&5\\
Portuguese&4991&58\\
Estonian&5076&14\\
\bottomrule
\end{tabular}
\caption{Number of lexemes (verbs, each consisting of different paradigm cells) and inflection classes in the used portion of the datasets.}
\label{tbl:datasets}
\end{table}

These three languages provide three distinct test cases for our model. Latin has a relatively regular inflection class system, governing affix variation only on a part of the paradigm cells, and is in this dataset only attested with macroclasses. This makes it an easy first test case for the model. Portuguese inflection classes determine affix variation across the whole paradigm, with a dataset containing microclasses, making it a more challenging test case. The system is also more opaque than Latin, because the theme vowel, which characterises classes, is visible in fewer paradigm cells than in Latin \citep[p. 5]{beniamine2021fine} Finally, Estonian has a system of inflectional classes defined by the distribution of stem variation within the paradigm, driven by a system of consonant gradation, making it a more challenging test case.

\subsection{Selection of cells}
\label{sec:selection}
Every lexeme is treated as a single data point in the clustering task. Such a data point combines all information on all the paradigm cells (different person-number-tense-mood combinations, such as present first singular) of the given lexeme. For this representation, a selection of cells is made that represent the inflectional system well and from which the algorithm should be able to infer the inflection classes. We do not use all paradigm cells (from a total of 50--70, depending on the language), because this makes running the algorithm too computationally expensive. For every language, we use a selection of cells, based on the information in the literature. 
Often, the full inflection of all cells for a verb, and thus the inflection class of the verb, can be predicted from just knowing a small number of cells (\emph{principal parts}; \citealp{finkel2007principal,finkel2009principal}). However, to model a more realistic learning task, we choose a larger selection of cells which gives a representative view of the whole paradigm. For some languages, we base this on a \emph{distillation}: when paradigm cells are grouped into sets such that a word form indicated by any cell within the set fully predicts the forms of all other cells in the sets (across a given verb), a distillation consists of one representative cell for every set \citep{pirrelli2000paradigmatic,bonami2003suppletion,stump2013morphological}. E.g. if knowing the \textsc{prs-ind.1sg} form of any verb always predicts the \textsc{prs-ind.3sg} form for the same verb, the set formed by them \{\textsc{prs-ind.1sg}, \textsc{prs-ind.3sg}\} can be distilled to only \{\textsc{prs-ind.1sg}\} (or \{\textsc{prs-ind.3sg}\} alternatively). Table \ref{tbl:cellselection} shows the selection of cells for the different languages.

\begin{table}
\begin{tabular}{lp{2.5cm}p{7cm}r}
\toprule
Language&Source&Cells&\#Cells\\
\midrule
Latin&\citet[Table 6]{pellegrini2020patterns} \newline (infectum and in dataset)&\textsc{imperf-ind.3sg}, \textsc{inf}, \textsc{imp.2sg}, \textsc{prs-ind.1sg}, \textsc{prs-ind.2sg}, \textsc{prs-ind.3sg}, \textsc{prs-ind.3pl}, \textsc{prs-sbjv.3sg}, \textsc{ger}&9\\
Portuguese&\citet[Table 18, p. 21]{beniamine2021fine}&\textsc{prs.ind.1sg}, \textsc{prs.ind.3sg}, \textsc{prs.ind.1pl}, \textsc{prs.ind.2pl}, \textsc{prs.ind.3pl}, \textsc{pst.impf.ind.3sg}, \textsc{pst.pfv.ind.1sg}, \textsc{pst.perf.ind.3sg}, \textsc{fut.ind.3sg}, \textsc{prs.sbjv.3sg}, \textsc{prs.sbjv.2pl}, \textsc{pst.ptcp}&12\\
Estonian&\citet[Table 4]{blevins2007conjugation}&\textsc{inf}, \textsc{imp.prs.2pl}, \textsc{imp.prs.pers}, \textsc{ger}, \textsc{ptcp.pst.pers}, \textsc{ind.prs.1sg}, \textsc{cond.prs.pers}, \textsc{imp.prs.2sg}, \textsc{sup}, \textsc{ptcp.prs.pers}, \textsc{quot.prs.pers}, \textsc{ind.pst.ipfv.1sg}, \textsc{ind.prs.impers}, \textsc{ind.pst.ipfv.impers}&14\\
\bottomrule
\end{tabular}
\caption{Selection of paradigm cells used for the different languages.}
\label{tbl:cellselection}
\end{table}

We base our selection of cells for Latin on the distillation used in \citet[Table 6, p. 212]{pellegrini2020patterns}. From the distilled cells, we only use those which belong to the infectum, because inflection classes only describe the variation of those cells. Furthermore, we omit cells that are not in our dataset: the future active, present passive infinitive and present participle. This leads us to a total of 9 cells for Latin, (see the first row in Table \ref{tbl:cellselection}).

In Portuguese, inflection classes describe the variation in the whole paradigm. We use the distillation from \citet[Table 18, p. 21]{beniamine2021fine}, totalling 12 cells.

Because inflection classes in Estonian describe distributional variation in the stem, rather than direct affix variation as seen in Latin and Portuguese, our selection of paradigm cells will also serve a different role: to capture the distribution of strong and weak stems within the paradigm to infer the inflection class. In principle, 1 form from the infinitive series and 1 form from the present series are enough to predict to which of the 3 macroclasses a verb belongs \citep[p. 262]{blevins2007conjugation}. To be able to predict the 14 microclasses in the dataset and make the learning task more realistic, we include a larger number of paradigm cells. We use the 14 cells from Table 4 from \citet{blevins2007conjugation} (also printed as Table \ref{tbl:estonianclasses-blevins2007} in this paper), which give a representative sample of cells from the 4 different series.

\subsection{Encoding}
\label{sub:encoding}
Phonemes occur in sequences, but this is difficult to represent in the type of Adaptive Resonance network that we use. Therefore, we use n-grams to capture part of the sequential information, in our case trigrams (cf. \citealp{baayen2019discriminative} for successful applications of trigrams in another computational modelling task).\footnote{See \citet{beniamine2018inferring} and \citet{beniamine2021multiple} for representations of forms which take into account segment ordering, by applying alignment.}\footnote{Trigrams capture a large amount of the sequential information: our target words can almost always be reconstructed on the basis of their trigrams.} As the ART1 model requires binary features as input, we only register presence or absence of trigrams. To create one datapoint per lexeme (e.g. \emph{stare} (lat.) `to stand' (eng.)), containing information on all cells (1\textsc{sg}, 2\textsc{sg}, ...), we create separate features registering trigram presence for a specific cell and concatenate all these features. This leads to a total of $n_{forms} \times n_{ngrams}$ features. 
Features which are 0 for every lexeme are removed, leading to removal of trigrams that do not exist for a cell across all lexemes.

In preliminary experiments, we also experimented with a set representation, in which a set of trigrams over the whole verb paradigm is taken (i.e. presence of an trigram occurring in multiple cells is only registered once). This representation gave, depending on the language, similar or sometimes even better results. However, as the goal of our model is to get better insight into human language processing, we only report results for the concatenation representation. This concatenation representation seems more cognitively plausible, as it allows for comparison of individual cells between different lexemes.

\subsection{Model}
As a model, we use Adaptive Resonance Theory 1 \citep{carpenter1987massively} (see section \ref{sec:art-background}). We used an implementation from the \texttt{neupy} library\footnote{The \texttt{neupy} library is still available at \url{https://github.com/itdxer/neupy}, but is no longer maintained.}, which we modified in several respects.\footnote{The modified code of the ART1 algorithm used for the experiments in this study, can be found in the GitHub repository of this paper: \texttt{https://github.com/peterdekker/agents-art}.} Firstly, we modified the algorithm to conform better to the specification of the ART1 algorithm: whereas the original \texttt{neupy} implementation used a fixed number of classes, we now start from 1 class and add classes dynamically. Secondly, we optimised the runtime of the code, specifically in the search procedure after the vigilance has not been overcome, by keeping an ordered list of candidate neurons based on bottom-up weights. For the learning rate $L$, we chose 2. The learning rate should always be higher than 1 \citep{britodasilva2019survey}, but we are not aware of work that empirically evaluates the influence of the learning rate value on model outcomes.

In our sample, the Latin dataset is augmented with macroclasses, while the Portuguese and Estonian datasets are attested with microclasses. The generalisation parameter (vigilance) in the ART1 neural network allows to adjust the model to the creation of smaller or larger clusters.
\subsection{Train-test regime}
The vigilance parameter in the network is one way to study generalisation in morphological processing: how different are different datapoints (lexemes) allowed to be in order to be assigned to the same cluster? In addition, we introduce a second way to test generalisation between lexemes. To see how well the model generalises to unseen data, and how well the clusters capture general knowledge of the inflectional system rather than idiosyncrasies of the observed data, we adopt a train-test regime. In this regime, common in supervised machine learning, the model is trained on one portion of the dataset and evaluated on another portion. We explain this regime in some detail, because for unsupervised methods, like the clustering approach proposed in this paper, the train-test regime is less common; in general, training and testing happen at the same time. However, in our clustering task, we do have a reference clustering (the inflection classes derived by linguists) to compare the clustering to, which is not always the case in unsupervised learning. This makes the train-test approach more relevant, as reliable data exists to evaluate the clustering during the two phases, instead of resorting to a measure of internal validity of the clusters.

For the test phase of the model, we developed an adjusted run of the ART1 algorithm, where datapoints are assigned to clusters, but no weights are updated. This testing functionality is not part of the original ART1 specification. The full ART1 algorithm was used, but the weight updates and the creation of new clusters were omitted. In the case where the match between a datapoint and a category template does not overcome the vigilance for any of the existing clusters and a new cluster would have to be created in the training phase, the datapoint is assigned to the category with the highest match instead.

\subsection{Evaluation}
To evaluate how well the model learns an inflection class system for different amounts of generalisation, we evaluate the models for vigilance values from 0.0 to 0.3, as prior experiments with our data shows that higher vigilance values create an unnecessarily high number (i.e. more than is necessary for accurate prediction of unseen word forms) of classes and drastically increase runtime of the algorithm. Per vigilance value (in steps of 0.005), 10 random permutations of the full dataset are created and a new model is created for every permutation. We plot 95\% confidence intervals over these runs.

Because this is an unsupervised algorithm where clusters do not have names that can be explicitly compared to targets, we cannot use traditional supervised metrics such as accuracy, precision and recall, but use a clustering similarity metric. The similarity between the clustering by the model and the linguistically attested inflection classes is evaluated using Adjusted Rand Index (ARI, \citealp{hubert1985comparing,steinley2004properties}). ARI is a metric that counts pairs of elements in the dataset that are assigned to the same cluster both in the model clustering and the reference clustering, which is adjusted for chance. ARI of 0 indicates random assignment to clusters and 1 indicates perfect clustering when compared to the reference clustering. The metric can also become negative for worse clustering than expected by chance \citep{chacon2023minimum}.

As a second similarity metric, for comparison, we report Adjusted Mutual Information (AMI; \citealp{vinh2010information}), the adjusted-for-chance version of Normalized Mutual Information, which measures similarity by comparing mutual information between the clusterings.\footnote{The unadjusted version, Normalized Mutual Information, with the arithmetic mean as averaging method, is equivalent to V-meaure, another clustering evaluation metric which is the harmonic mean between homogeneity and completeness \citep{rosenberg2007vmeasure}.} In preliminary experiments, we found that ARI penalises too high numbers of clusters (number of clusters > number of attested inflection classes) better than AMI. Therefore, we use ARI as the leading metric and report AMI for comparison.

As a simple baseline model to compare ART1 clustering performance, we also report ARI and AMI for a \emph{k-means} model, an oft-used clustering algorithm which tries to minimise a criterion based on the sum of squares of the distance of the assigned datapoints to the cluster centre (see \citealp{sculley2010webscale}). The algorithm requires setting the number of clusters beforehand, which we set to the number of inflection classes in the dataset. Implementations of ARI, AMI and k-means from the \texttt{scikit-learn} library were used \citep{pedregosa2011scikitlearn}.

To analyse the behaviour of the model, we will also report the attested inflection classes of the assigned lexemes per cluster (as a bar plot) and the distinctive trigrams per cluster, extracted from the templates in the top-down weights (see Section \ref{sec:art-background}).

\section{Results}
We performed inflection class clustering using ART1 for three languages with different characteristics, in increasing difficulty: Latin, Portuguese and Estonian and evaluated how well the clustering corresponds to attested inflection classes. We also report how well the clustering algorithm generalises to a held-out test set for the three languages.

\subsection{Latin: Testing model capabilities}
The relatively small dataset for Latin, the regularity of the Latin verbal inflection system and the annotation of the dataset with only a few macroclasses, allow for a first test case to evaluate the capabilities of the ART1 model on the inflection class clustering task.

\begin{figure}
\begin{subfigure}[T]{0.49\linewidth}
    \centering
    \includegraphics[width=\linewidth,trim={0cm 0cm 1.75cm 1.5cm},clip]{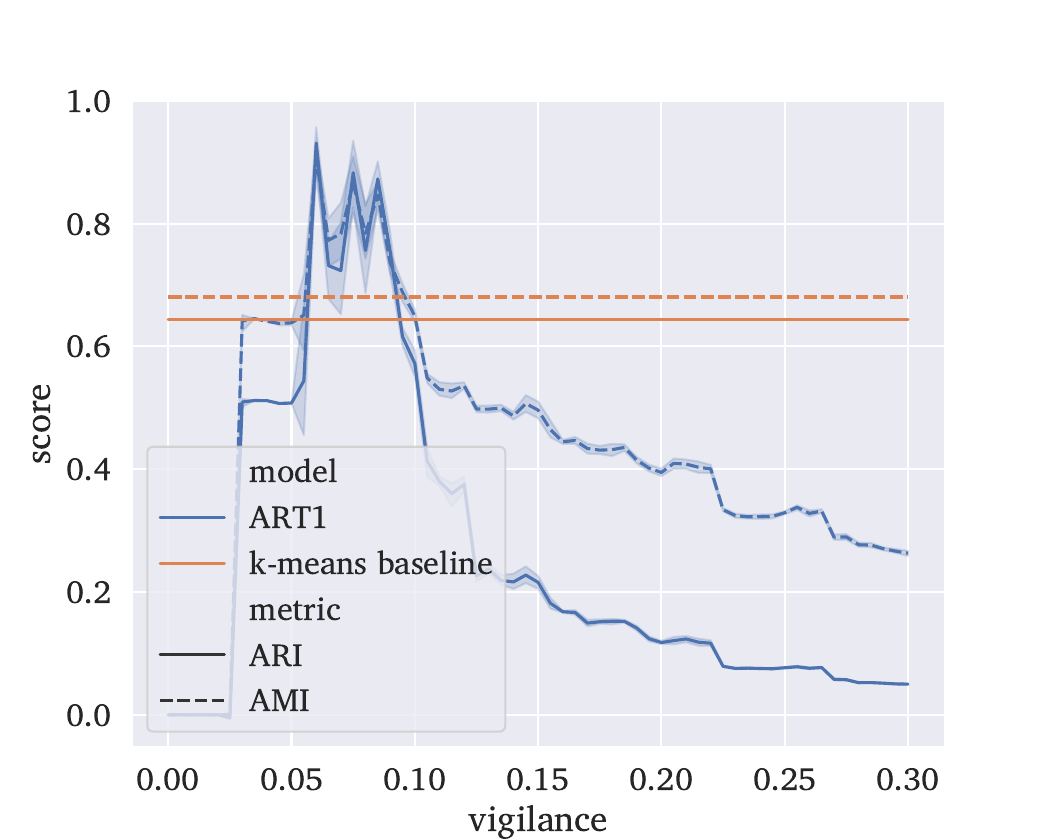} 
    \caption{Similarity to attested inflection classes}
    \label{fig:latin-distill-concat-ari}
\end{subfigure}
\begin{subfigure}[T]{0.49\linewidth}
    \centering
    \includegraphics[width=\linewidth,trim={0cm 0cm 1.75cm 1.5cm},clip]{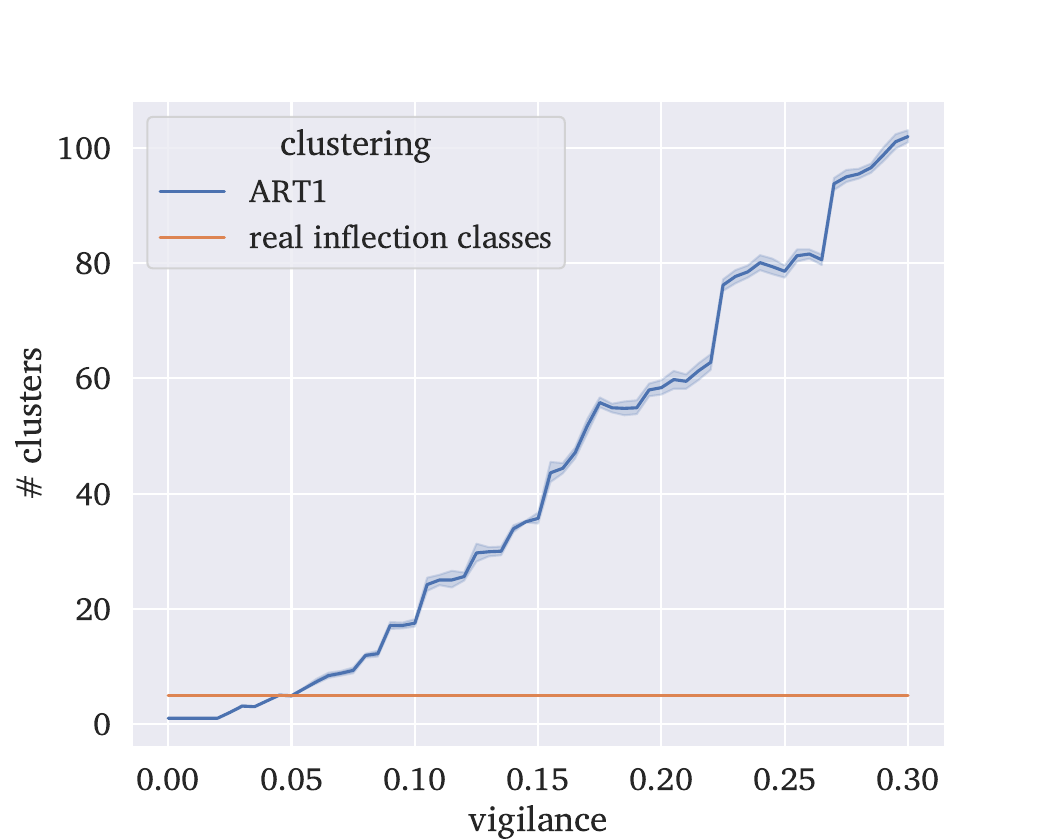} 
    \caption{Number of clusters\\(line: real \#inflection classes)}
    \label{fig:latin-distill-concat-clusters}
\end{subfigure}
\caption{Results ART1 for Latin for different vigilance values (95\% confidence intervals over 10 random permutations of data). Left figure: solid line = AMI (Adjusted Mutual Information), dashed line = ARI (Adjusted Rand Index).}
\label{fig:viglatin-distill}
\end{figure}

Figure \ref{fig:viglatin-distill} shows the results of the model for Latin: the similarity to the linguistically attested inflection classes, measured in ARI and AMI (Figure \ref{fig:latin-distill-concat-ari}) and the number of clusters (\ref{fig:latin-distill-concat-clusters}), to be able to compare to the real number of clusters in the dataset. For Latin, the clustering matches the attested inflection classes very well, ARI is 0.94 for vigilance 0.06. For this vigilance value, the number of clusters (7) is also close to the number of inflection classes proposed by linguists (5). In general, the ART1 model is able to cluster well in a narrow vigilance range between 0.05 and 0.010, for which it also outperforms the k-means baseline.

\begin{figure}
    \includegraphics[width=\linewidth]{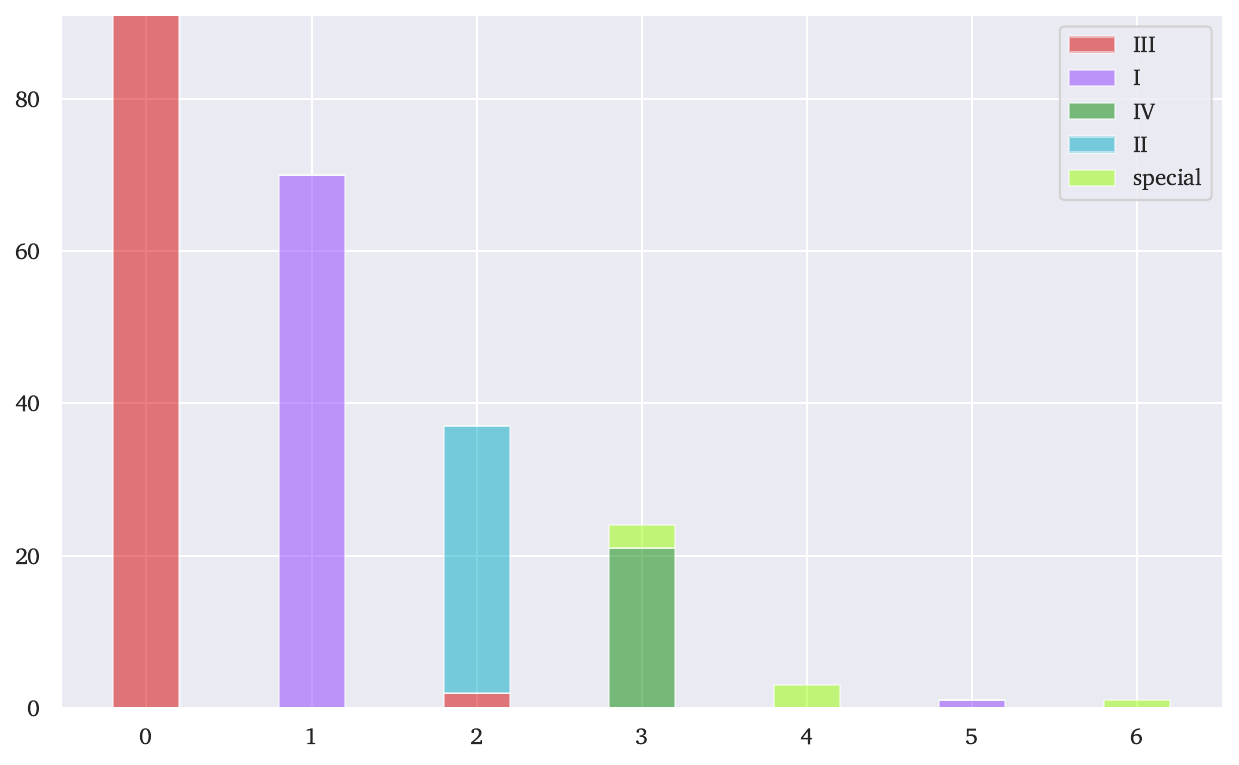} 
    \caption{Assigned lexemes per cluster, single run ART1 for Latin (vigilance 0.06). Bar = cluster, colour = attested inﬂection class of assigned lexemes.}
    \label{fig:clusters-latin}
\end{figure}

Figure \ref{fig:clusters-latin} gives an analysis of the clustering for the highest-performing vigilance value of 0.060. Figure  \ref{fig:clusters-latin} shows a bar for every discovered cluster, while the colours signify the attested inflection classes of the lexemes that have been assigned to the clusters.  We can observe that the clustering matches the attested inflection classes well: many clusters almost exclusively contain lexemes from one attested inflection class. However, the number of clusters is larger than the number of linguistic classes. Some classes (notably \emph{special}) have been spread over multiple clusters, but this is partly because we do not make an effort to filter out irregular (unpredictable) forms. The ``excess'' clusters 5 and 6 have attracted very few datapoints.

Table \ref{fig:ngrams-latin} shows the trigrams for every cluster (with the same cluster numbering as in Figure \ref{fig:clusters-latin}). As discussed in Section \ref{sec:art-background}, the top-down weights of the network consist of one template per cluster, which represents the features (trigrams per paradigm cell) the network attends to for that cluster. Following every trigram feature, we report the proportion of the total number of lexemes with this feature that are assigned to this cluster: a higher proportion means that the feature is more unique for this cluster and occurs in fewer other clusters. Behind the vertical bar, we give the total number of lexemes with this feature. For comparison, the attested inflection class of the majority of the lexemes assigned to the cluster is also indicated in the table (which corresponds to colour filling most of the bar in Figure \ref{fig:clusters-latin}).

We can observe that the trigrams per cluster match well with the \emph{theme vowels} traditionally associated with inflection classes in Latin. For example cluster 1 shows the \emph{a} theme vowel characteristic of class I and cluster 2 shows the \emph{e} theme vowel characteristic of class II. Unsurprisingly, class 4 has different trigrams which cannot easily be matched to a certain type of verb, as the irregular verbs in class \emph{special} are quite different from each other.

\begin{table}
\tiny
\begin{tabular}{lp{0.7\linewidth}p{0.2\linewidth}}
\toprule
cluster & distinctive trigrams & majority class \\
\midrule
0 & \textsc{inf}: \textit{ere} (1.000 | 91)\newline\textsc{prs-ind.3pl}: \textit{unt} (0.958 | 95)\newline\textsc{ger}: \textit{end} (0.599 | 152), \textit{ndoː} (0.401 | 227)\newline\textsc{imperf-ind.3sg}: \textit{eːba} (0.599 | 152), \textit{bat} (0.401 | 227) & III \\
1 & \textsc{ger}: \textit{and} (1.000 | 70), \textit{ndoː} (0.308 | 227)\newline\textsc{imperf-ind.3sg}: \textit{aːba} (1.000 | 70), \textit{bat} (0.308 | 227)\newline\textsc{prs-ind.3pl}: \textit{ant} (1.000 | 70)\newline\textsc{inf}: \textit{aːre} (1.000 | 70) & I \\
2 & \textsc{inf}: \textit{eːre} (1.000 | 37)\newline\textsc{prs-ind.3pl}: \textit{ent} (1.000 | 37)\newline\textsc{prs-sbjv.3sg}: \textit{eat} (0.974 | 38)\newline\textsc{ger}: \textit{end} (0.243 | 152), \textit{ndoː} (0.163 | 227)\newline\textsc{imperf-ind.3sg}: \textit{eːba} (0.243 | 152), \textit{bat} (0.163 | 227) & II \\
3 & \textsc{imperf-ind.3sg}: \textit{eːba} (0.158 | 152), \textit{bat} (0.106 | 227)\newline\textsc{ger}: \textit{end} (0.158 | 152), \textit{ndoː} (0.106 | 227) & IV \\
4 & \textsc{prs-ind.3pl}: \textit{unt} (0.032 | 95)\newline\textsc{imperf-ind.3sg}: \textit{bat} (0.013 | 227)\newline\textsc{ger}: \textit{ndoː} (0.013 | 227) & special \\
5 & \textsc{prs-sbjv.3sg}: \textit{aːns} (1.000 | 1), \textit{raːn} (1.000 | 1), \textit{traː} (1.000 | 1), \textit{sea} (1.000 | 1), \textit{nse} (1.000 | 1), \textit{eat} (0.026 | 38)\newline\textsc{prs-ind.3pl}: \textit{aːns} (1.000 | 1), \textit{raːn} (1.000 | 1), \textit{eun} (1.000 | 1), \textit{seu} (1.000 | 1), \textit{nse} (1.000 | 1), \textit{traː} (1.000 | 1), \textit{unt} (0.011 | 95)\newline\textsc{imperf-ind.3sg}: \textit{traː} (1.000 | 1), \textit{aːns} (1.000 | 1), \textit{raːn} (1.000 | 1), \textit{siːb} (1.000 | 1), \textit{iːba} (1.000 | 1), \textit{nsiː} (1.000 | 1), \textit{bat} (0.004 | 227)\newline\textsc{imp.2sg}: \textit{aːns} (1.000 | 1), \textit{traː} (1.000 | 1), \textit{nsiː} (1.000 | 1), \textit{raːn} (1.000 | 1)\newline\textsc{prs-ind.1sg}: \textit{raːn} (1.000 | 1), \textit{seoː} (1.000 | 1), \textit{traː} (1.000 | 1), \textit{aːns} (1.000 | 1), \textit{nse} (1.000 | 1)\newline\textsc{ger}: \textit{und} (1.000 | 1), \textit{nse} (1.000 | 1), \textit{aːns} (1.000 | 1), \textit{traː} (1.000 | 1), \textit{eun} (1.000 | 1), \textit{seu} (1.000 | 1), \textit{raːn} (1.000 | 1), \textit{ndoː} (0.004 | 227)\newline\textsc{prs-ind.3sg}: \textit{raːn} (1.000 | 1), \textit{nsi} (1.000 | 1), \textit{aːns} (1.000 | 1), \textit{sit} (1.000 | 1), \textit{traː} (1.000 | 1)\newline\textsc{inf}: \textit{siːr} (1.000 | 1), \textit{aːns} (1.000 | 1), \textit{iːre} (1.000 | 1), \textit{raːn} (1.000 | 1), \textit{nsiː} (1.000 | 1), \textit{traː} (1.000 | 1)\newline\textsc{prs-ind.2sg}: \textit{raːn} (1.000 | 1), \textit{traː} (1.000 | 1), \textit{siːs} (1.000 | 1), \textit{nsiː} (1.000 | 1), \textit{aːns} (1.000 | 1) & I \\
6 & \textsc{imperf-ind.3sg}: \textit{bat} (0.004 | 227)\newline\textsc{ger}: \textit{ndoː} (0.004 | 227) & special \\
\bottomrule
\end{tabular}
    \caption{Trigrams per cluster, single run ART1 for Latin (vigilance 0.06). In brackets: proportion of total number of lexemes with this feature that are assigned to this cluster (higher is more unique) and total number of lexemes with this feature}
    \label{fig:ngrams-latin}
\end{table}

\subsection{Portuguese: Modelling micro-variation}
\label{sec:resultsportuguese}
Because the results for Latin showed that the ART1 model can perform a clustering into macroclasses, we now turn to a dataset of Portuguese, which is annotated with a larger number of more detailed microclasses. In this way, we can test if the model is powerful enough to model this finer-grained variation and cluster lexemes into microclasses. In Portuguese, the theme vowels are only visible in some paradigm cells, while in other cells, they are neutralised to be the same vowel for all inflection classes \citep[p. 22]{beniamine2021fine}. This makes it more difficult to infer the inflection class of a lexeme in Portuguese than in Latin. In addition, the vowel preceding the theme vowel may alternate across paradigm cells, based on the stress position in that paradigm cell \citep[p. 24-28]{beniamine2021fine}. Even though this prethematic vowel is not diagnostic of the inflection class (the theme vowel defines the inflection class), this adds extra unpredictability to the whole system

\begin{figure}
\begin{subfigure}[T]{0.49\linewidth}
    \centering
    \includegraphics[width=\linewidth,trim={0cm 0cm 1.75cm 1.5cm},clip]{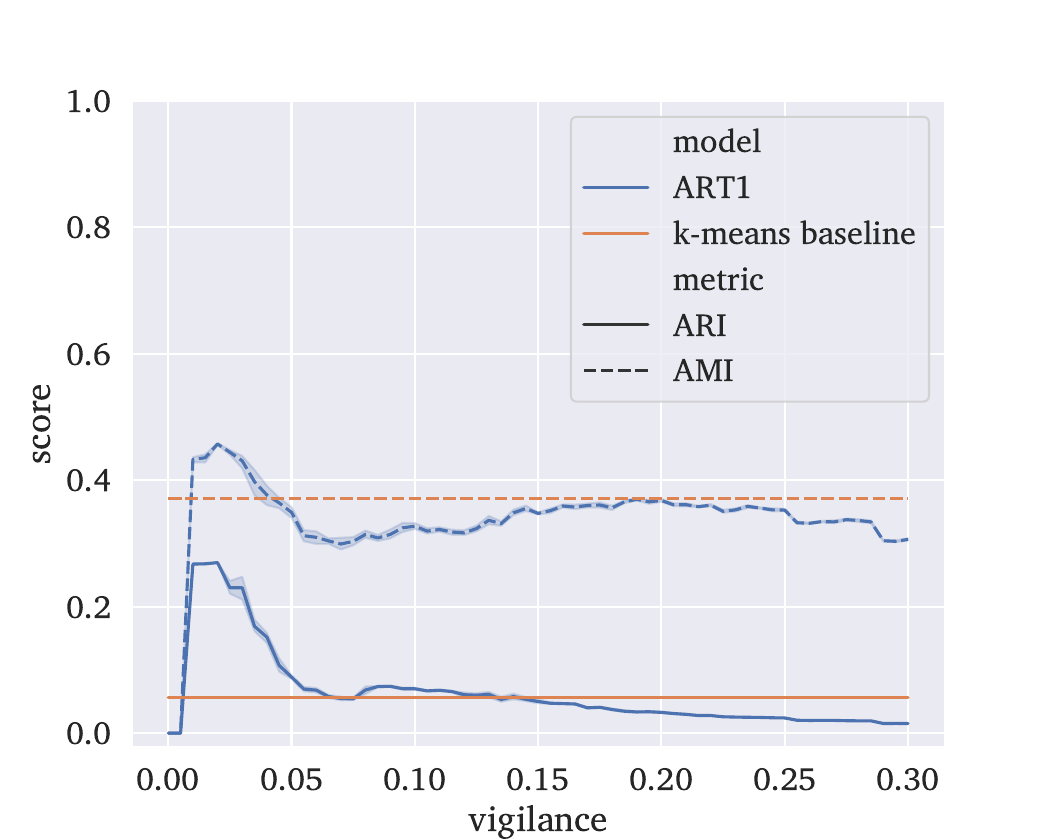} 
    \caption{Similarity to attested inflection classes\\}
    \label{fig:portuguese-distill-concat-ari}
\end{subfigure}
\begin{subfigure}[T]{0.49\linewidth}
    \centering
    \includegraphics[width=\linewidth,trim={0cm 0cm 1.75cm 1.5cm},clip]{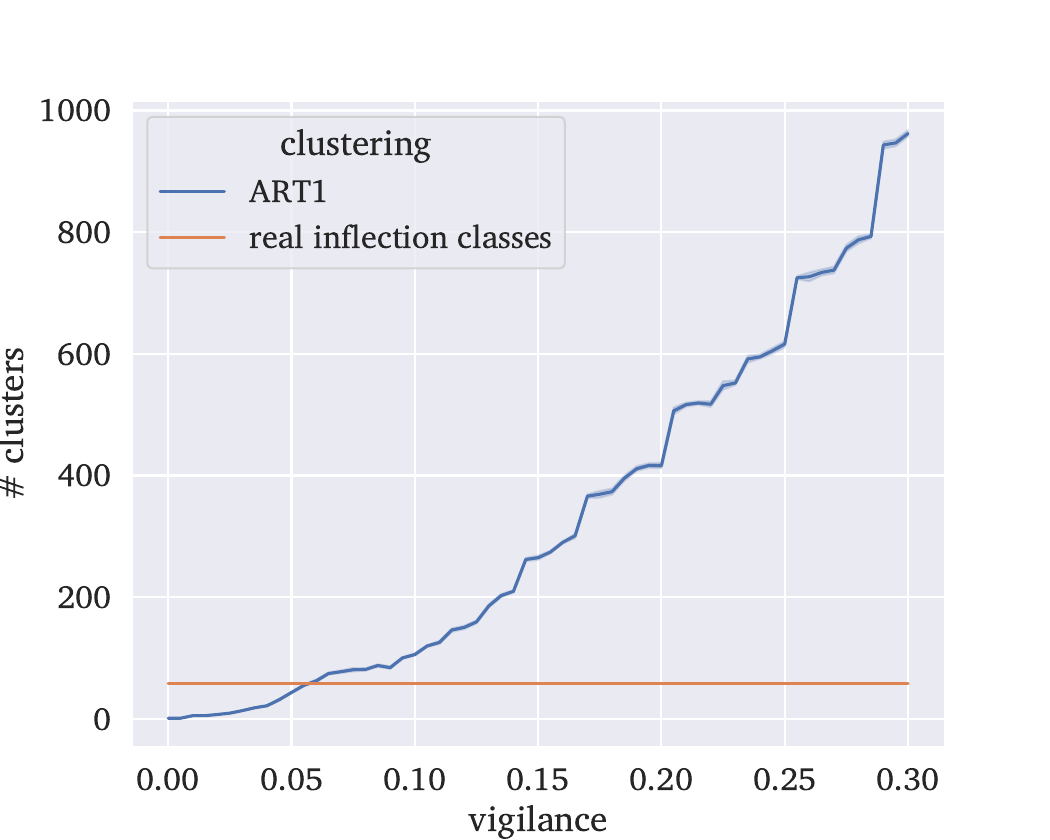} 
    \caption{Number of clusters\\(line: real \#inflection classes)}
    \label{fig:portuguese-distill-concat-clusters}
\end{subfigure}
\caption{Results ART1 for Portuguese for different vigilance values (95\% confidence intervals over 10 random permutations of data). Left figure: solid line = AMI (Adjusted Mutual Information), dashed line = ARI (Adjusted Rand Index).}
\label{fig:vigportuguese-distill}
\end{figure}

Figure \ref{fig:vigportuguese-distill} shows the model results for Portuguese. Clustering similarity here is lower than for Latin, with the highest ARI value around 0.27 for a vigilance value of 0.020. However, around these vigilance values, the number of clusters is 5-10, much lower than the number of cluster labels proposed by the linguists who annotated this dataset (58).

\begin{figure}
    \includegraphics[width=\linewidth]{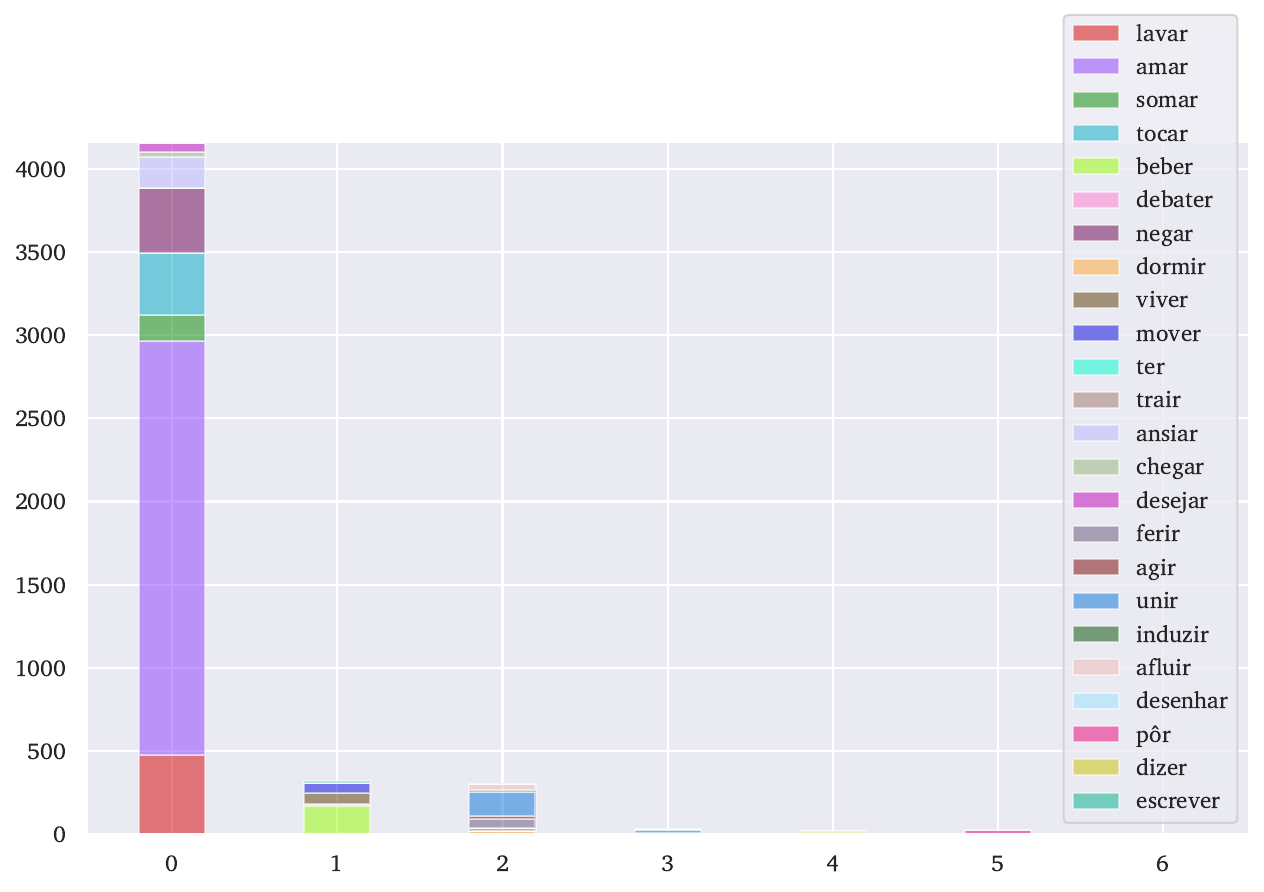} 
    \caption{Assigned lexemes per cluster, single run ART1 for Portuguese (vigilance 0.02). Bar = cluster, colour = attested inﬂection class of assigned lexemes. Only showing inflection classes with more than 10 lexemes.}
    \label{fig:clusters-portuguese}
\end{figure}

To analyse the clusters, we use the best-performing vigilance value of 0.020. For visibility in the plot and table, because of the large number of inflection classes in Portuguese, 
we only showed the lexemes (coloured segments in the bars) for attested inflection classes in the dataset that have 10 or more lexemes: datapoints for attested inflection classes with fewer lexemes are not shown, the bars are thus lower. Figure \ref{fig:clusters-portuguese} shows that the model creates a few large clusters and after that many smaller clusters. The clusters contain lexemes from multiple inflection classes. However, it does seem that the first three clusters cluster together macroclasses with the same theme vowel and in this way represent the three traditional macroclasses \citep[p. 4]{beniamine2021fine}: \emph{-ar} (cluster 0), \emph{-er} (cluster 1) and \emph{-ir} (cluster 3). Table \ref{fig:ngrams-portuguese} shows the trigrams per cluster, which confirms that the first three classes learned the theme vowels \emph{a}, \emph{e} and \emph{i}. Clusters 3--6 have attracted very few datapoints (in this plot somewhat reinforced by only showing inflection classes with more than 10 lexemes). Cluster 5 is a small cluster that seems to coincide perfectly with the (small) microclass \emph{pôr}: this class consists of 29 lexemes, and indeed it can be observed in the numbers behind the features that all 29 lexemes which have this feature have been assigned to this cluster. The model is thus able to cluster lexemes into macroclasses and also recognises specific microclasses (such as \emph{pôr}) well.

\begin{table}
\begin{tabular}{lp{0.7\linewidth}p{0.2\linewidth}}
\toprule
cluster & distinctive trigrams & majority class \\
\midrule
0 & \textsc{prs.ind.1pl}: \textit{ˈɐmu} (1.000 | 4168), \textit{muʃ} (0.835 | 4991)\newline\textsc{pst.impf.ind.3sg}: \textit{ˈavɐ} (1.000 | 4168)\newline\textsc{fut.ind.3sg}: \textit{ɐɾˈa} (1.000 | 4168) & amar \\
1 & \textsc{fut.ind.3sg}: \textit{əɾˈa} (1.000 | 350)\newline\textsc{pst.perf.ind.3sg}: \textit{ˈeɾɐ} (1.000 | 350)\newline\textsc{prs.ind.1pl}: \textit{ˈemu} (0.884 | 396), \textit{muʃ} (0.070 | 4991) & beber \\
2 & \textsc{prs.ind.1pl}: \textit{ˈimu} (1.000 | 344), \textit{muʃ} (0.069 | 4991)\newline\textsc{pst.perf.ind.3sg}: \textit{ˈiɾɐ} (1.000 | 344)\newline\textsc{fut.ind.3sg}: \textit{iɾˈa} (0.866 | 397) & unir \\
3 & \textsc{fut.ind.3sg}: \textit{iɾˈa} (0.134 | 397)\newline\textsc{prs.ind.1pl}: \textit{muʃ} (0.011 | 4991) & unir \\
4 & \textsc{prs.ind.1pl}: \textit{ˈemu} (0.116 | 396), \textit{muʃ} (0.009 | 4991) & ter \\
5 & \textsc{prs.ind.1sg}: \textit{ˈoɲu} (1.000 | 29), \textit{pˈoɲ} (1.000 | 29)\newline\textsc{pst.impf.ind.3sg}: \textit{pˈuɲ} (1.000 | 29), \textit{ˈuɲɐ} (1.000 | 29)\newline\textsc{prs.sbjv.3sg}: \textit{pˈoɲ} (1.000 | 29), \textit{ˈoɲɐ} (1.000 | 29)\newline\textsc{fut.ind.3sg}: \textit{uɾˈa} (1.000 | 29), \textit{puɾ} (1.000 | 29)\newline\textsc{prs.sbjv.2pl}: \textit{ɲˈajʃ} (1.000 | 29), \textit{puɲ} (1.000 | 29), \textit{uɲˈaj} (1.000 | 29)\newline\textsc{pst.perf.ind.3sg}: \textit{ˈɛɾɐ} (1.000 | 29), \textit{puz} (1.000 | 29), \textit{zˈɛɾ} (1.000 | 29), \textit{uzˈɛ} (1.000 | 29)\newline\textsc{pst.pfv.ind.1sg}: \textit{pˈuʃ} (1.000 | 29)\newline\textsc{prs.ind.3pl}: \textit{pˈõjɐ̃j} (1.000 | 29)\newline\textsc{prs.ind.2pl}: \textit{ˈõdə} (1.000 | 29), \textit{dəʃ} (1.000 | 29), \textit{pˈõd} (1.000 | 29)\newline\textsc{prs.ind.1pl}: \textit{pˈom} (1.000 | 29), \textit{ˈomu} (1.000 | 29), \textit{muʃ} (0.006 | 4991) & pôr \\
6 & \textsc{prs.ind.1pl}: \textit{muʃ} (0.000 | 4991) & ser \\
\bottomrule
\end{tabular}
    \caption{Trigrams per cluster, single run ART1 for Portuguese (vigilance 0.02). In brackets: proportion of total number of lexemes with this feature that are assigned to this cluster (higher is more unique) and total number of lexemes with this feature.}
    \label{fig:ngrams-portuguese}
\end{table}

\subsection{Estonian: Learning distributed patterns}
\label{sec:resultsestonian}
The results of the Latin dataset evaluate performance on macroclasses while the results of the Portuguese dataset evaluate performance on microclasses. However, both languages have inflection classes that describe affixal variation. Now we turn to Estonian, for which the dataset is also annotated with microclasses. However, in Estonian, inflection classes describe variation in stems rather than affixes. Moreover, the inflection classes are not determined by a concrete sequence of segments in the stem, but based on the distribution of the morphophonological property of consonant gradation over paradigm cells (see Section \ref{sec:selection}). Finding inflection classes for Estonian seems more difficult than for the other languages, for two reasons. Firstly, because the strong and weak grade are relative patterns: \emph{kk} is the strong grade of weak \emph{k}, but \emph{k} can also be the strong grade paired with weak \emph{∅}. In order to learn such a system, it would seem necessary that the model can learn alternation patterns within the paradigm instead of looking only at the surface strings. Secondly, inflection classes in Estonian are defined by a distribution over cells. For this, a model has to learn logical relations between cells within one verb paradigm (datapoint), rather than just comparing similarity between datapoints.

\begin{figure}
\begin{subfigure}[T]{0.49\linewidth}
    \centering
    \includegraphics[width=\linewidth,trim={0cm 0cm 1.75cm 1.5cm},clip]{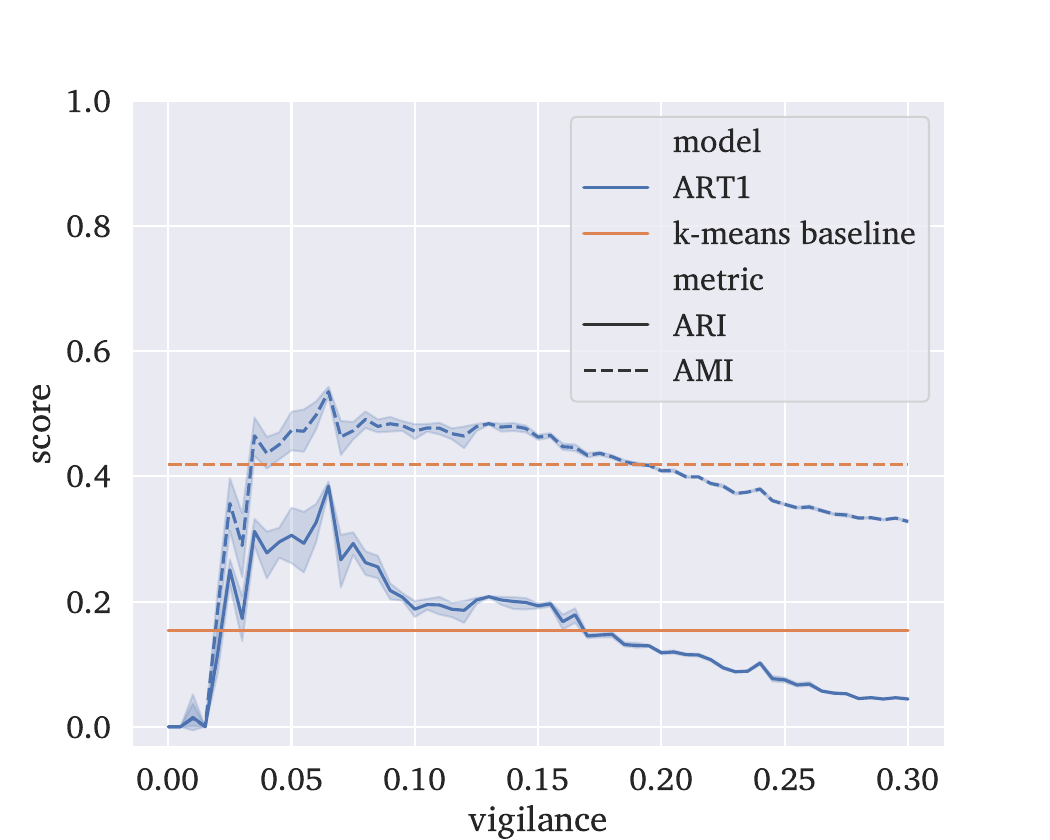} 
    \caption{Similarity to attested inflection classes\\}
    \label{fig:estonian-distill-concat-ari}
\end{subfigure}
\begin{subfigure}[T]{0.49\linewidth}
    \centering
    \includegraphics[width=\linewidth,trim={0cm 0cm 1.75cm 1.5cm},clip]{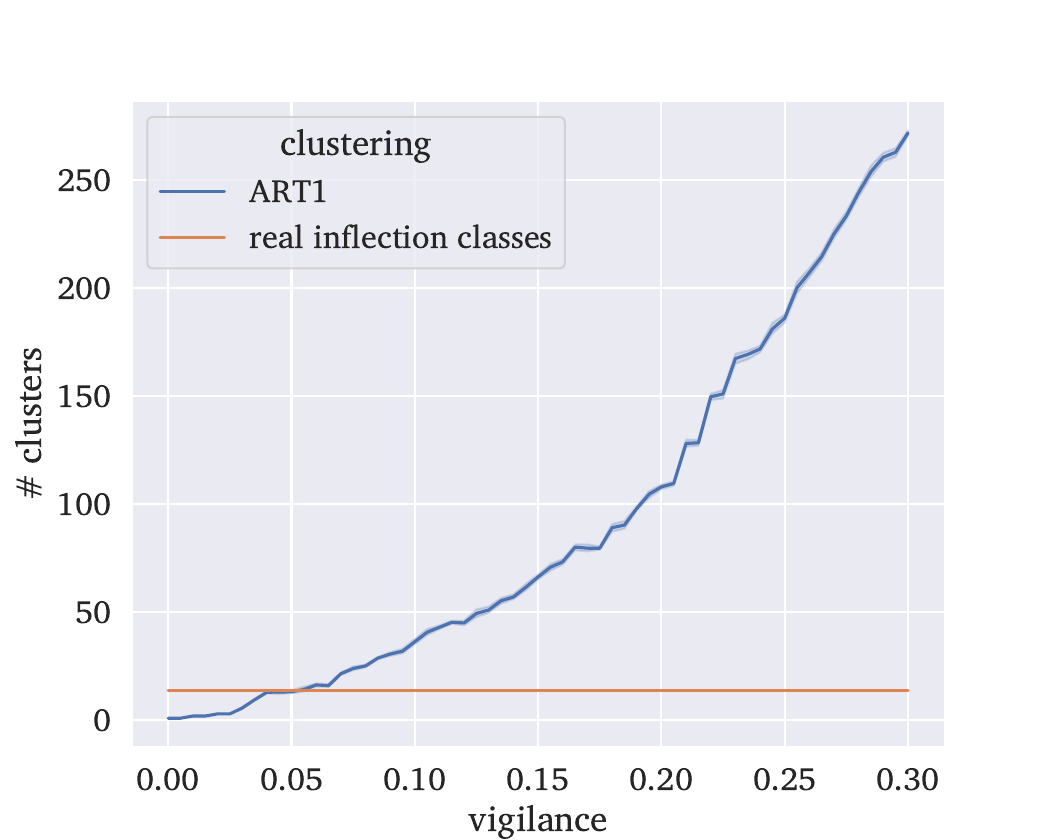} 
    \caption{Number of clusters\\(line: real \#inflection classes)}
    \label{fig:estonian-distill-clusters}
\end{subfigure}
\caption{Results ART1 for Estonian for different vigilance values (95\% confidence intervals over 10 random permutations of data). Left figure: solid line = AMI (Adjusted Mutual Information), dashed line = ARI (Adjusted Rand Index).}
\label{fig:vigestonian-distill}
\end{figure}

Figure \ref{fig:vigestonian-distill} shows results for Estonian. The similarity to the linguistically attested inflection classes is highest for vigilance 0.060, with an ARI of almost 0.4. This is significantly lower than the performance for Latin, but higher than the result for Portuguese, even though the at first sight, the Estonian conjugation system seemed more complex than that of Portuguese. The ART1 model clearly outperforms the k-means baseline.  For the vigilance value of 0.040, the number of clusters is also similar to the number of inflection classes described by linguists (Figure \ref{fig:estonian-distill-clusters}). Possibly the smaller number of inflection classes (14 in Estonian versus 58 in Portuguese) makes the task easier, but clearly the model is able to infer inflection classes without explicit information about consonant gradation.

\begin{figure}
    \includegraphics[width=\linewidth]{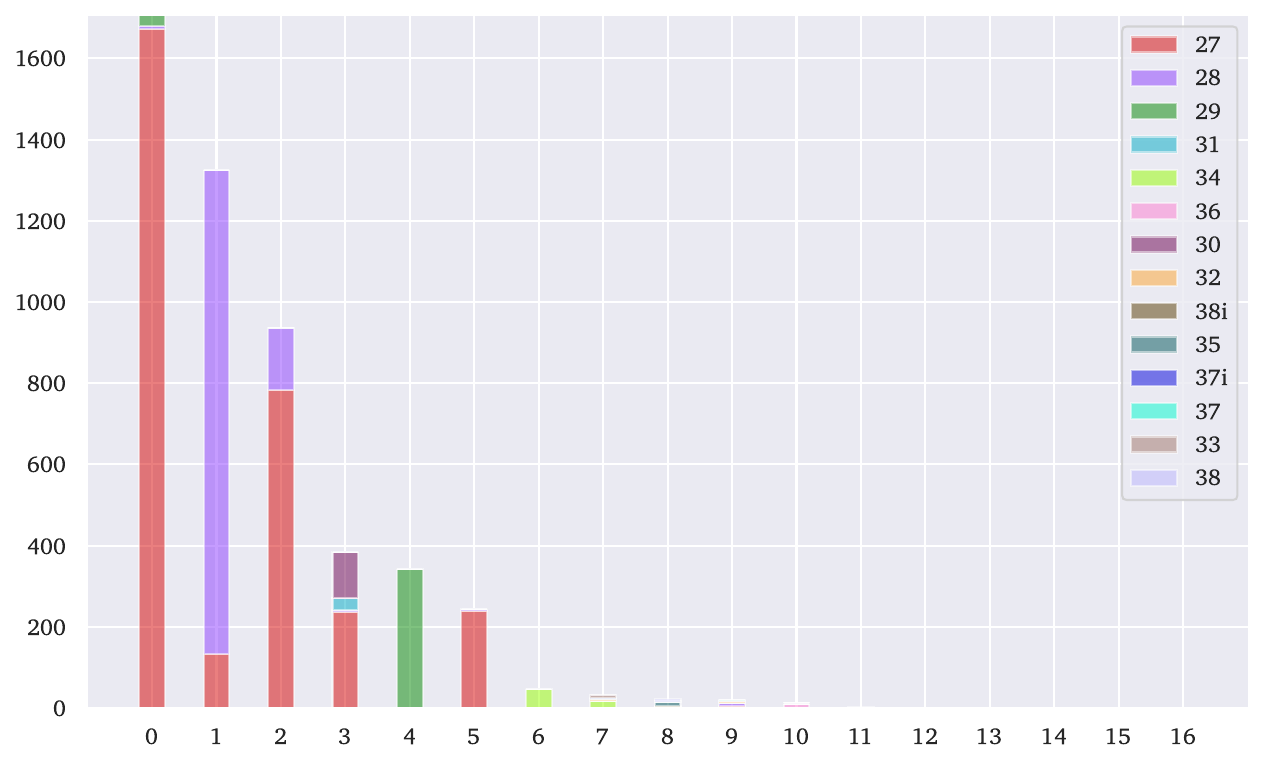} 
    \caption{Assigned lexemes per cluster, single run ART1 for Estonian (vigilance 0.06). Bar = cluster, colour = attested inﬂection class of assigned lexemes.}
    \label{fig:clusters-estonian}
\end{figure}

The assignment of lexemes to inflection classes (Figure \ref{fig:clusters-estonian}) for the highest-performing vigilance value of 0.060 shows that most clusters for Estonian are relatively pure: they contain lexemes from only a few inflection classes recognized by linguists. On the other hand, lexemes belonging to one inflection class are often spread out over multiple clusters. This spread-out effect could possibly be caused by the distributed phonological patterns in Estonian: because one surface string can be both the weak grade of one form and the strong grade of another form, lexemes from one inflection class can get assigned to multiple clusters. Clusters 7 -- 16 have attracted very few datapoints and are almost unused. Table \ref{fig:ngrams-estonian} shows the distinctive trigrams per cluster in Estonian (for the largest 8 clusters only for visibility). The strong/weak alternations characteristic of the inflection classes (such as \textit{pp}/\textit{p}; Table \ref{tbl:estonianclasses-blevins2007}) cannot easily be discerned here: this may indeed indicate that the model is not able to capture distributional patterns of the paradigm beyond those that can be inferred by looking at the surface strings, as discussed above. Instead, it seems the algorithm has clustered verbs together with the same vowels in the stem, such as \emph{a} in cluster 0, \emph{i} in cluster 1 and \emph{u} in cluster 2. Stem vowels can in some cases give cues to infer about which of the three macroclasses a verb belongs to, but is not a reliable indication \citep[p. 255--262]{blevins2007conjugation}.

\begin{table}
\tiny
\begin{tabular}{lp{0.7\linewidth}p{0.2\linewidth}}
\toprule
cluster & distinctive trigrams & majority class \\
\midrule
0 & \textsc{ger}: \textit{ɑte} (1.000 | 1707), \textit{tes} (0.367 | 4649)\newline\textsc{inf}: \textit{ɑtɑ} (1.000 | 1707)\newline\textsc{imp.prs.pers}: \textit{ɑku} (1.000 | 1707)\newline\textsc{imp.prs.2pl}: \textit{ɑke} (1.000 | 1707)\newline\textsc{ind.pst.ipfv.1sg}: \textit{ɑsʲi} (0.833 | 2050), \textit{sʲin} (0.344 | 4962)\newline\textsc{quot.prs.pers}: \textit{ɑvɑ} (0.832 | 2051), \textit{vɑtː} (0.336 | 5076)\newline\textsc{ptcp.pst.pers}: \textit{ɑnu} (0.832 | 2051), \textit{nut} (0.336 | 5076)\newline\textsc{sup}: \textit{ɑmɑ} (0.832 | 2051)\newline\textsc{ind.prs.impers}: \textit{kse} (0.336 | 5075), \textit{ɑks} (0.336 | 5075) & 27 \\
1 & \textsc{imp.prs.2pl}: \textit{ike} (0.999 | 1327)\newline\textsc{inf}: \textit{itɑ} (0.999 | 1327)\newline\textsc{quot.prs.pers}: \textit{ivɑ} (0.999 | 1327), \textit{vɑtː} (0.261 | 5076)\newline\textsc{imp.prs.pers}: \textit{iku} (0.999 | 1327)\newline\textsc{ger}: \textit{ite} (0.999 | 1327), \textit{tes} (0.285 | 4649)\newline\textsc{sup}: \textit{imɑ} (0.999 | 1327)\newline\textsc{ind.pst.ipfv.1sg}: \textit{isʲi} (0.999 | 1327), \textit{sʲin} (0.267 | 4962)\newline\textsc{ptcp.pst.pers}: \textit{inu} (0.999 | 1327), \textit{nut} (0.261 | 5076)\newline\textsc{ind.prs.impers}: \textit{ɑks} (0.261 | 5075), \textit{kse} (0.261 | 5075) & 28 \\
2 & \textsc{imp.prs.2pl}: \textit{uke} (1.000 | 936)\newline\textsc{quot.prs.pers}: \textit{uvɑ} (1.000 | 936), \textit{vɑtː} (0.184 | 5076)\newline\textsc{sup}: \textit{umɑ} (1.000 | 936)\newline\textsc{imp.prs.pers}: \textit{uku} (1.000 | 936)\newline\textsc{ind.pst.ipfv.1sg}: \textit{usʲi} (1.000 | 936), \textit{sʲin} (0.189 | 4962)\newline\textsc{inf}: \textit{utɑ} (1.000 | 936)\newline\textsc{ptcp.pst.pers}: \textit{unu} (1.000 | 936), \textit{nut} (0.184 | 5076)\newline\textsc{ger}: \textit{ute} (1.000 | 936), \textit{tes} (0.201 | 4649)\newline\textsc{ind.prs.impers}: \textit{ɑks} (0.184 | 5075), \textit{kse} (0.184 | 5075) & 27 \\
3 & \textsc{ind.pst.ipfv.1sg}: \textit{esʲi} (0.612 | 629), \textit{sʲin} (0.078 | 4962)\newline\textsc{cond.prs.pers}: \textit{eks} (0.601 | 641)\newline\textsc{quot.prs.pers}: \textit{evɑ} (0.601 | 641), \textit{vɑtː} (0.076 | 5076)\newline\textsc{sup}: \textit{emɑ} (0.601 | 641)\newline\textsc{ger}: \textit{tes} (0.083 | 4649)\newline\textsc{ind.prs.impers}: \textit{ɑks} (0.076 | 5075), \textit{kse} (0.076 | 5075)\newline\textsc{ptcp.pst.pers}: \textit{nut} (0.076 | 5076) & 27 \\
4 & \textsc{cond.prs.pers}: \textit{ɑks} (1.000 | 343)\newline\textsc{ind.pst.ipfv.1sg}: \textit{ɑsʲi} (0.167 | 2050), \textit{sʲin} (0.069 | 4962)\newline\textsc{ptcp.pst.pers}: \textit{ɑnu} (0.167 | 2051), \textit{nut} (0.068 | 5076)\newline\textsc{quot.prs.pers}: \textit{ɑvɑ} (0.167 | 2051), \textit{vɑtː} (0.068 | 5076)\newline\textsc{sup}: \textit{ɑmɑ} (0.167 | 2051)\newline\textsc{ind.prs.impers}: \textit{ɑks} (0.068 | 5075), \textit{kse} (0.068 | 5075) & 29 \\
5 & \textsc{ptcp.pst.pers}: \textit{enu} (1.000 | 244), \textit{nut} (0.048 | 5076)\newline\textsc{ger}: \textit{ete} (1.000 | 244), \textit{tes} (0.052 | 4649)\newline\textsc{imp.prs.2pl}: \textit{eke} (1.000 | 244)\newline\textsc{imp.prs.pers}: \textit{eku} (1.000 | 244)\newline\textsc{inf}: \textit{etɑ} (1.000 | 244)\newline\textsc{ind.pst.ipfv.impers}: \textit{etʲːi} (0.838 | 291)\newline\textsc{ind.prs.impers}: \textit{etːɑ} (0.838 | 291), \textit{tːɑk} (0.833 | 293), \textit{kse} (0.048 | 5075), \textit{ɑks} (0.048 | 5075)\newline\textsc{ind.pst.ipfv.1sg}: \textit{esʲi} (0.388 | 629), \textit{sʲin} (0.049 | 4962)\newline\textsc{cond.prs.pers}: \textit{eks} (0.381 | 641)\newline\textsc{sup}: \textit{emɑ} (0.381 | 641)\newline\textsc{quot.prs.pers}: \textit{evɑ} (0.381 | 641), \textit{vɑtː} (0.048 | 5076) & 27 \\
6 & \textsc{ind.pst.ipfv.impers}: \textit{etʲːi} (0.162 | 291)\newline\textsc{ind.prs.impers}: \textit{etːɑ} (0.162 | 291), \textit{tːɑk} (0.160 | 293), \textit{kse} (0.009 | 5075), \textit{ɑks} (0.009 | 5075)\newline\textsc{quot.prs.pers}: \textit{vɑtː} (0.009 | 5076)\newline\textsc{ptcp.pst.pers}: \textit{nut} (0.009 | 5076) & 34 \\
7 & \textsc{ger}: \textit{tes} (0.007 | 4649)\newline\textsc{ind.prs.impers}: \textit{ɑks} (0.006 | 5075), \textit{kse} (0.006 | 5075)\newline\textsc{quot.prs.pers}: \textit{vɑtː} (0.006 | 5076)\newline\textsc{ptcp.pst.pers}: \textit{nut} (0.006 | 5076) & 34 \\
\end{tabular}
    \caption{Trigrams per cluster, single run ART1 for Estonian (vigilance 0.06), for largest 8 clusters. In brackets: proportion of total number of lexemes with this feature that are assigned to this cluster (higher is more unique) and total number of lexemes with this feature}
    \label{fig:ngrams-estonian}
\end{table}

\subsection{Generalising to unseen data}
\label{sec:generalisation}
The results show that clusters are learned that are similar to the way linguists cluster words in inflection classes. However, the precise clustering depends on the value of the vigilance parameter. These values fall in a limited range (0.02--0.06) but the optimal value is different for the three languages. This can be understood as follows: when clustering stimuli (in this case words) into classes (in this case inflection classes) languages determine membership of clusters in different ways. This is also observed in other domains where languages cluster stimuli, for instance, when clustering colour stimuli into colour terms. In our study, this is modelled by the value of the vigilance parameter. In real language learning, the value of the clustering parameters would be determined by feedback on how well the learned clusters allow a speaker to correctly generalize to unobserved word forms. In this study, we will not go so far as to try to determine the appropriate value of the vigilance parameter automatically, but it is interesting to verify whether the learned clusters allow us to generalize to unseen forms correctly.

\begin{figure}
\begin{subfigure}[b]{0.32\linewidth}
    \centering
    \includegraphics[width=\linewidth,trim={0cm 0cm 1.25cm 1.5cm},clip]{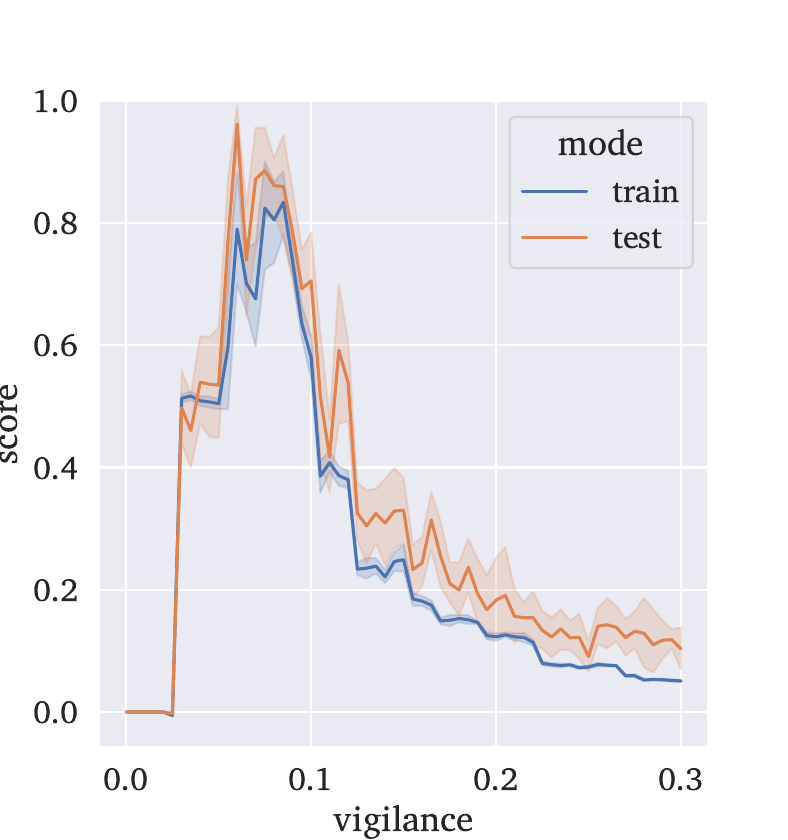} 
    \caption{Latin}
    \label{fig:traintestlatin}
\end{subfigure}
\begin{subfigure}[b]{0.32\linewidth}
    \centering
    \includegraphics[width=\linewidth,trim={0cm 0cm 1.25cm 1.5cm},clip]{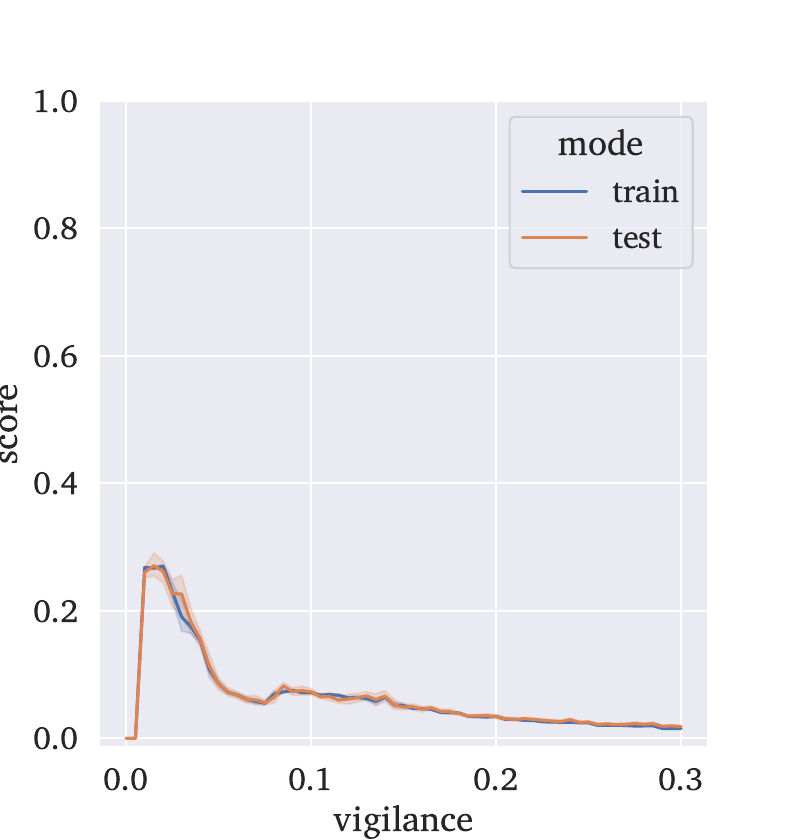} 
    \caption{Portuguese}
    \label{fig:traintestportuguese}
\end{subfigure}
\begin{subfigure}[b]{0.32\linewidth}
    \centering
    \includegraphics[width=\linewidth,trim={0cm 0cm 1.25cm 1.5cm},clip]{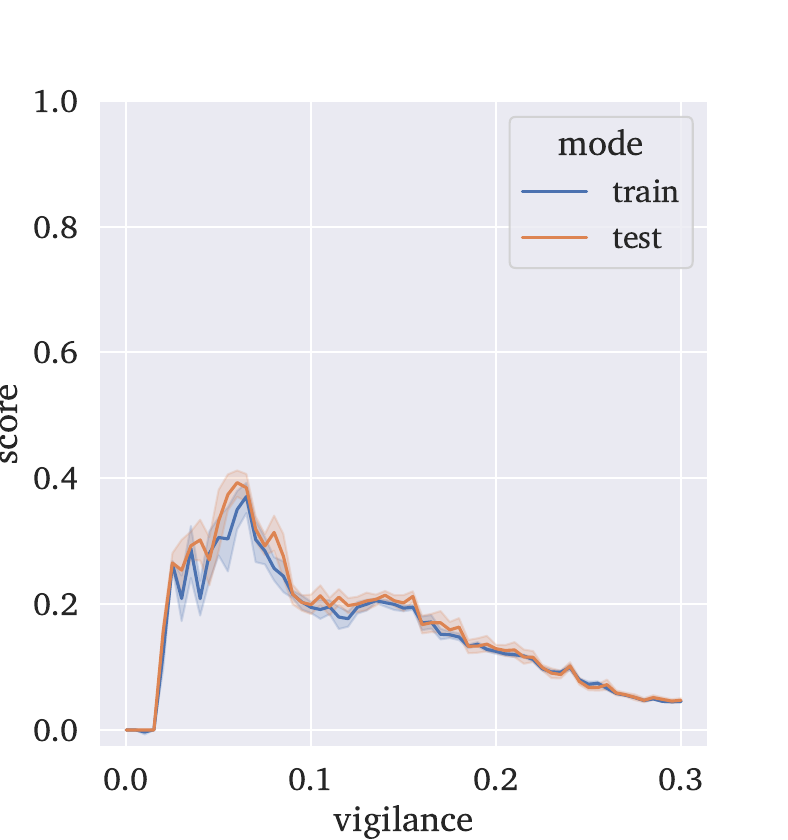} 
    \caption{Estonian}
    \label{fig:traintestestonian}
\end{subfigure}
\caption{Results of ART1 on train-test generalisation (ARI score), 10-fold cross validation (90\% train, 10\% test).}
\label{fig:traintest}
\end{figure}

We therefore performed train-test runs for Latin, Estonian and Portuguese, with 90\% train data and 10\% test data. 10-fold cross-validation was used to make sure performance is consistent over different splits of the data. In this paradigm, data is split into 10 so-called \emph{folds}, and in 10 different runs, the data is divided into a train set of 9 folds and a test set of 1 fold, where folds are alternated. Results are shown in Figure \ref{fig:traintest}. It can be observed that for all the languages, clustering similarity for the test set, measured in ARI, is similar or even higher than that of the training set. This means that the model learned good generalisations on the training set that can be applied to the test set.

\section{Towards a diachronic model}
Our experiments have shown that inflection classes can be learned from surface information and that the classes can be used to predict unseen morphological forms. This provides a first step towards modelling the role of inflection classes in language change. In order to flesh this out a little further, it is worthwhile to think about how our model could be extended to a diachronic simulation of language change.
Several agent-based models of the evolution of inflectional morphology have been proposed, in which agents communicate using word forms \citep{hare1995learning,cotterell2018diachronic}. A number of these studies specifically investigate the evolution of inflection classes \citep{ackerman2015no,parker2018bayesian,round2021role,round2022cognition,round2023role}. In these models, the change over time of inflection classes is the object of study, but in most models, inflection classes are not explicitly part of the communicative task of the agents.

In order to extend our model to the case of language change, it is necessary to include the inflection class clustering task and the ART model in an agent-based model in which agents transmit linguistic information to each other. Models from the ART family have been used in multi-agent models in which agents can act upon their environment, \citep{tan2005selforganizing,xiao2007selforganizing} but the ART1 model alone is not sufficient for studying communication or language transmission. It is a clustering model, but for communication which entails both comprehension and (re-) production of language, a way is needed to reproduce linguistic forms from the learned clusters. We would suggest to not only equip every agent with an ART1 model, performing the task of inflection class clustering, but also to equip them with the model to reproduce seen and unseen forms from the learned clusters. We could for instance use forms derived in analogy to the most prevalent forms in our clusters. Our results show that even for a language such as Estonian, for which the clustering task is hard, the resulting clusters are relatively pure, i.e. one inflection class according to the linguistic analysis dominates each cluster (figure \ref{fig:clusters-estonian}). Moreover section \ref{sec:generalisation} shows that unseen froms are clustered similarly to previously seen word forms. Agents could then interact with other agents in a paradigm that \citet{steels1995selforganizing} has called the \emph{language game}. 

In a typical language game, agents interact in pairs, exchanging some kind of linguistic information and learning from the interaction. In our case, agents would produce inflected forms of words. Their production would be based on a form of the Paradigm Cell Filling Problem (as in the model in \citealp{round2022cognition}). Inflection class clustering is then  a separate model component, which provides a `prior' for the main component that solves the Paradigm Cell Filling Problem. The clustering into inflection classes could provide agents with information about which verbs are inflected in the same way. In this way we could study transmission from generation to generation using a form of \emph{iterated learning} \citep{kirby2002emergence,kirby2014iterated}, i.e. agents from the next generation learn on the basis of utterances produced by agents from the previous generation. Moreover, if the agents would be initialised with word forms without a developed inflection class system, such experiments could also be used to study the initial emergence of inflection classes (cf. \citealp{round2021role,round2023role}).

\section{Conclusion and Discussion}
To study how learnable, and thus how prone to change, a system of inflection classes is, we proposed the task of clustering lexemes into inflection classes, with an Adaptive Resonance Theory 1 neural network as a computational model of morphological processing. We showed that ART1, using a simple trigram representation of the phonemic wordforms, is able to learn the relatively regular and coarse-grained system of verbal inflection classes of Latin very well. For Portuguese and Estonian, performance of the model is reasonable, but some aspects of the morphology seem fundamentally too difficult to learn in the current combination of model, task setting and data representation. However, we did find (Figure \ref{fig:traintest}) that generalisations to unseen word forms appeared to be equally good for all languages -- i.e. that word forms from the unseen test set were clustered equally well as word forms that appeared in the training set. Apparently our model, which Grossberg \citep{grossberg1976adaptive, grossberg2020path} has based on cognitive and neurological principles, can learn inflection classes reasonably well, based on only phonemic (surface) forms (the results for our different languages are discussed in more detail below). 

Reasonably well in this context means that the learned clusters can be used to predict inflections for unseen forms. Although there is no one-to-one correspondence between linguistically derived inflection classes and our clusters, the clusters tend to attract lexemes that are inflected in the same way. We can therefore use the cluster in which an unseen form falls to predict how it inflects. This is somewhat surprising, because the features we use are not fine-tuned for this task at all, and ART uses simple conjunction (logical \textsc{AND}) of features to assign an input to a cluster. Still, precise clustering depends on the vigilance parameter, and its optimal value depends on the language, indicating that this also needs to be learned. This might however be possible from interactions with other speakers.

Because of the interpretability of the ART1 model via its cluster templates, we are able to assess the features that the model learned. This shows that, for Latin and Portuguese, the model indeed learns the patterns in the stem characteristic of the inflection classes, while for Estonian the model learned patterns (stem vowels) that co-variate with inflection classes but do not fully define them.

Whereas a part of the performance of the model for the different languages can be explained by the capabilities of the model, the chosen representation of the data also plays an important role. As a starting point, a representation was used where a data point corresponds to a collection of trigrams of the paradigm cells for one lexeme. This representation could in the future be enriched to contain more linguistic information or replaced by a more advanced representation, for example using aligned word forms (see \citealp{beniamine2018inferring,beniamine2021multiple}). However, it would be desirable that this is still based as much as possible on directly observable (surface) information. We will now examine the results for different languages, to identify for which aspects a different encoding of the data would be helpful and which aspects constitute fundamental limitations of the chosen ART model and clustering setting.

As discussed in Section \ref{sec:resultsportuguese}, the Portuguese system is more difficult to predict than that of Latin, because the theme vowel, diagnostic of the inflection class, is not visible in all paradigm cells. A data point for a lexeme, which contains information on multiple cells in the paradigm, thus contains a weaker signal of the inflection class, with more divergent other information, than in for example Latin, where most cells bear a theme vowel. Additionally, the vowel preceding the theme vowel varies across paradigm cells based on the stress position. Although this prethematic vowel does not define the inflection class, it creates extra noise in the data and possibly leads to clustering based on spurious similarity between data points. The model performance for Portuguese is indeed much lower than for Latin, which makes it plausible that these two sources of noise have played a role. In general, one way to improve the data representation would be to add stress markers to the phonetic forms, so a model can discern that prethematic vowel variation co-occurs with stress. However, for the current data representation, with multiple paradigm cells per data point and features separately registered per paradigm cell, this seems less useful. Similarity between data points (lexemes) is determined by comparing the same paradigm cells in the respective lexemes, so co-variation of stress position and vowel alternation across paradigm cells is not taken into account. In a different task setting, where one paradigm cell has to be predicted from another and a data point corresponds to an individual paradigm cell, such stress marking would be more useful.

For the Estonian verbal morphology system, two difficulties were identified (see Section \ref{sec:resultsestonian}): the context-dependency of consonant gradation (the same sound can be the weak grade of one verb and the strong grade of another) and the distributed patterns of cells defining inflection classes. The first problem could be addressed by choosing a representation where forms are explicitly coded for the weak or strong grade\footnote{This representation could be something along the lines of what \citet{lefevre2021formalizing} calls \emph{de-identification}, replacing concrete phonemes by more abstract variables.}, although adding this linguistic information makes the task less natural. The second problem is more fundamental: ART1 works based on similarity between data points and does not take into account the relationships between features (cells) within a data point (verb paradigm). Table \ref{tbl:estonianclasses-blevins2007} illustrates why similarity between data points cannot fully capture the distributional system in Estonian: based on similarity, lexemes from class I and II would be closer together than lexemes from class I and III, because they share more cells with the same grade, while in fact there is no ground for this difference. To overcome this problem within the inflection class clustering task, we need to be able to capture relations between the forms for a single lexeme, something which is not possible in the present representation. We would need a different representation of the data, where one data point corresponds to one paradigm cell rather than a full paradigm. The challenge would then be to explicitly model inflection classes in this task. A potential idea is to use feature transformations (based on deep learning) that are learned simultaneously with the clustering. Transformers \citep{vaswani2017attention} are a technique from deep learning that has been used successfully for learning language, and although these models tend to be extremely complex, it is entirely possible to create a small and simple instance of a transformer for our model. 

One of the contributions of the Adaptive Resonance Theory model is that the level of generalisation is a parameter of the model, which allows evaluation of different generalisation policies and therefore provides insights into the possible cognitive processes needed for processing of inflection classes. The models performed best for relatively low vigilance values for the different languages (0.02-0.06). However, the best-performing vigilance value depends on the task, the number of classes in the data, and the chosen representation. With a different data representation, data points may have shown less or more similarity, making it easier or harder for the match between the datapoint and the category template to overcome the vigilance parameter. In a more complete system (for instance, the agent-based model that we proposed for investigating language change) it is therefore necessary to automatically tune the vigilance parameter, based on whether a learner is successful in reproducing the correct morphological forms. This seems feasible as there is only a narrow peak of vigilance values for which the algorithm performs well (and there is low performance outside that peak) in all three languages.

As one of the goals of this paper was to shed light on change of inflection classes, we sketched a possible approach to simulate evolution of inflection classes by creating an agent-based simulation with ART networks as models of language processing of the individual agents. We hope that this can lead to a line of work where cognitively plausible models, such as ART1, are applied to get a better understanding of morphological processing both synchronically and diachronically.

\section{Acknowledgements}
All code for this project can be found on Zenodo as \url{https://zenodo.org/records/13356366} and is available in the GitHub repository \url{https://github.com/peterdekker/agents-art}.

This work was supported by funding from the Flemish Government under the \emph{Onderzoeksprogramma Artificiële Intelligentie (AI) Vlaanderen} programme. PD was supported by a PhD Fellowship fundamental research (11A2821N) of the \emph{Research Foundation -- Flanders (FWO)} and HR was supported by a Senior Postdoctoral Fellowship (1258822N) of the \emph{Research Foundation -- Flanders (FWO)}.

\bibliographystyle{apacite}
\bibliography{art}

\end{document}